\documentclass{article}
\usepackage[final]{colm2026_conference}

\usepackage{microtype}
\usepackage{hyperref}
\usepackage{url}
\usepackage{booktabs}
\usepackage{lineno}
\usepackage{array}
\usepackage{amsmath}
\usepackage{tikz}
\usepackage{subcaption}
\usepackage{wrapfig}
\usepackage{fontawesome5}
\usepackage[most]{tcolorbox}
\usetikzlibrary{arrows.meta, decorations.pathreplacing, shapes.geometric}

\newtcolorbox{examplebox}[1][]{
  enhanced,
  colback=blue!3!white,
  colframe=blue!45!black,
  arc=2mm,
  boxrule=0.4pt,
  left=4pt, right=4pt, top=4pt, bottom=4pt,
  breakable,
  #1
}

\definecolor{darkblue}{rgb}{0, 0, 0.5}
\definecolor{clusA}{RGB}{230,140,40}
\definecolor{clusB}{RGB}{60,160,90}
\definecolor{clusC}{RGB}{60,120,200}
\definecolor{cellOK}{RGB}{120,180,90}
\definecolor{cellOld}{RGB}{220,90,90}
\definecolor{cellMix}{RGB}{245,200,80}
\hypersetup{colorlinks=true, citecolor=darkblue, linkcolor=darkblue, urlcolor=darkblue}

\usepackage{fvextra}
\usepackage{inconsolata}
\usepackage{placeins}
\newcommand{\ehl}[2]{{\setlength{\fboxsep}{1pt}\colorbox{#1}{#2}}}
\newcommand{\eA}{\ehl{clusA!22}{A}}
\newcommand{\eB}{\ehl{clusB!22}{B}}
\newcommand{\eC}{\ehl{clusC!22}{C}}

\title{Set-shifting Behavioral Test for Harnessed Agents}

\author{Ye Ziwei\\
\texttt{yeziwei.ml@gmail.com}}

\begin{document}

\ifcolmsubmission
\linenumbers
\fi

\maketitle
\suppressfloats[t]

\raggedbottom

\begin{abstract}
  What happens to an LLM agent's tool choice when the reliable tool silently changes within an ongoing session? We borrow the notion of set-shifting from cognitive psychology to study how well agents adapt to hidden reliability shifts. Our cognitive test for LLM agents mounts libraries of redundant tools and skills, in which many tools solve the same task but differ in hidden reliability. Using a branching schedule, we shift the reliable tool group in the environment and compare it with a stable control, allowing us to isolate the effect of each shift on the agent's behavior. We conduct our study on a panel of LLMs equipped with harnesses and show that the same set of shifts results in distinct behaviors across models: some latch onto a fixed routine within a few turns, whereas others continue to vary. Less capable models often omit the reliable tool group, while frontier models keep calling it alongside the other groups. We introduce a suite of measures to quantify agent behavior after reliability shifts. While policy prompting substantially alters behavior in some tested models, our findings highlight agents' brittleness when changes occur in indirectly observable context.
\end{abstract}

\section{Introduction}

Humans who repeatedly solve similar problems form habits that simplify future decisions but can persist past their usefulness. The Einstellung effect captures how people persist with a learned solution when a better one is available \citep{luchins1942einstellung}. The Wisconsin Card Sorting Test (WCST) makes the challenge more difficult by changing the sorting rule without warning mid-task \citep{grant1948wcst,heaton1993wcst}. Meanwhile, reversal learning paradigms apply hidden changes to value-based choice \citep{izquierdo2017reversal}. Recent work ports these cognitive tests to LLMs in textual or visual sorting tasks \citep{li2024reflectionbench,hao2025vlmwcst,li2025nearoptimal}. 

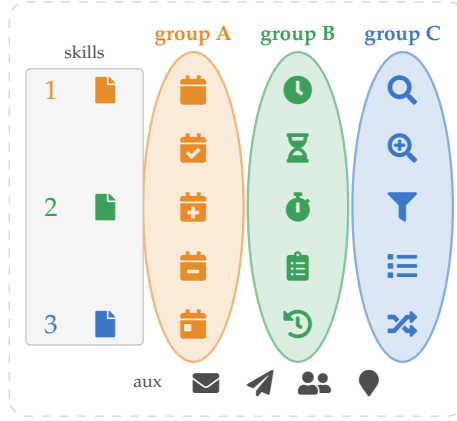
\begin{figure}[t]
\centering
\resizebox{0.44\textwidth}{!}{%
\begin{tikzpicture}[
  font=\small,
  clusblob/.style={ellipse, draw=#1!60, fill=#1!18, minimum width=12mm, minimum height=37mm, thick},
  tool/.style={text=#1, font=\normalsize}
]
  \def\xS{0.85} \def\xA{2.15} \def\xB{3.4} \def\xC{4.65}
  \def\rA{2.95} \def\rB{2.25} \def\rC{1.55} \def\rD{0.85} \def\rE{0.15}

  \node[clusblob=clusA] at (\xA, 1.55) {};
  \node[clusblob=clusB] at (\xB, 1.55) {};
  \node[clusblob=clusC] at (\xC, 1.55) {};

  \node[font=\scriptsize\bfseries, text=clusA] at (\xA, 3.6) {group A};
  \node[font=\scriptsize\bfseries, text=clusB] at (\xB, 3.6) {group B};
  \node[font=\scriptsize\bfseries, text=clusC] at (\xC, 3.6) {group C};

  \node[tool=clusA] at (\xA, \rA) {\faCalendar};
  \node[tool=clusA] at (\xA, \rB) {\faCalendarCheck};
  \node[tool=clusA] at (\xA, \rC) {\faCalendarPlus};
  \node[tool=clusA] at (\xA, \rD) {\faCalendarMinus};
  \node[tool=clusA] at (\xA, \rE) {\faCalendarDay};

  \node[tool=clusB] at (\xB, \rA) {\faClock};
  \node[tool=clusB] at (\xB, \rB) {\faHourglassHalf};
  \node[tool=clusB] at (\xB, \rC) {\faStopwatch};
  \node[tool=clusB] at (\xB, \rD) {\faClipboardList};
  \node[tool=clusB] at (\xB, \rE) {\faHistory};

  \node[tool=clusC] at (\xC, \rA) {\faSearch};
  \node[tool=clusC] at (\xC, \rB) {\faSearchPlus};
  \node[tool=clusC] at (\xC, \rC) {\faFilter};
  \node[tool=clusC] at (\xC, \rD) {\faList};
  \node[tool=clusC] at (\xC, \rE) {\faRandom};

  \node[rounded corners=1.5pt, draw=gray!50, fill=gray!8,
        minimum width=14mm, minimum height=33mm] at (\xS, 1.55) {};
  \node[font=\tiny, gray!60!black] at (\xS, 3.4) {skills};
  \node[font=\footnotesize, text=clusA] at (\xS-0.4, \rA) {1};
  \node[font=\small, text=clusA]        at (\xS+0.25, \rA) {\faFile};
  \node[font=\footnotesize, text=clusB] at (\xS-0.4, \rC) {2};
  \node[font=\small, text=clusB]        at (\xS+0.25, \rC) {\faFile};
  \node[font=\footnotesize, text=clusC] at (\xS-0.4, \rE) {3};
  \node[font=\small, text=clusC]        at (\xS+0.25, \rE) {\faFile};

  \node[font=\tiny, gray!60!black, anchor=east] at (1.9, -0.55) {aux};
  \node[gray!60!black, font=\footnotesize] at (2.3, -0.55) {\faEnvelope};
  \node[gray!60!black, font=\footnotesize] at (2.95, -0.55) {\faPaperPlane};
  \node[gray!60!black, font=\footnotesize] at (3.6, -0.55) {\faUserFriends};
  \node[gray!60!black, font=\footnotesize] at (4.25, -0.55) {\faMapMarker};

  \draw[gray!35, dashed, thin, rounded corners=6pt]
    (-0.05,-0.95) rectangle (5.4,4.0);
\end{tikzpicture}
}
\caption{\textbf{Tool-skill structure.} Each SKILL.md (left) describes one tool group, and each group contains functionally redundant variants. Boundaries silently change which group is reliable.}
\label{fig:ecology}
\end{figure}

Modern LLM agents reside in agentic harnesses in which each call to the model arrives with prior context. These segmented context blocks include tool calls and results, skills and tool schemas, and system prompts, so the agent never starts from a blank slate. In such a rich environment, a mounted library of skills and tools may contain several nominally equivalent capabilities, each calling different APIs and backend services. As with humans, repeated tool use recorded within the harness context lets the agent converge on a preferred subset of skills and tools. An agent's tool preference may outlive an unannounced backend change, leaving the agent using failing toolsets. To test agent set-shifting, we ask: Can an agent revise its action route when the reliable tool group changes silently?

\begin{table*}[t]
\begin{center}
\small
\begin{tabular}{p{0.18\linewidth}p{0.38\linewidth}p{0.30\linewidth}}
\toprule
\textbf{Set Type} & \textbf{Manipulation} & \textbf{Example} \\
\midrule
Description variant & Same schema and execution, different name and phrasing & \texttt{calendar\_lookup} vs. \texttt{slot\_window\_lookup} \\
Quality gradient & Same visible intent, different hidden reliability across phases & tool group reliable in phase 1 vs. unreliable in phase 2 \\
Side effect conflict & Mutating action vs. reversible preview & \texttt{send\_email} vs. \texttt{draft\_email} \\
\bottomrule
\end{tabular}
\end{center}
\caption{Tool-skill set relation types.}
\label{tab:conflict-types}
\end{table*}

We make the following contributions: (i) A behavioral test for harnessed agents based on \emph{redundant tool-skill sets}\footnote{Code: \url{https://github.com/zwycl/agent-wcst-mini-bench}}, libraries of many functionally equivalent tools where the working subset silently shifts mid-trajectory. (ii) A descriptive framework that characterizes post-shift behavior as two routing profiles, concentration and retention. (iii) A case study applying our experimental protocol to a panel of both open-weight and closed-weight LLMs, with a structural \emph{set framing} intervention and an instructional \emph{policy} applied across the panel. Together, these contributions point to a capability-linked pattern in behavior (Figure~\ref{fig:intel}).

\section{Behavioral Test}

\subsection{Tool-Skill Sets and Taxonomy}

\paragraph{Definitions.} In the agentic harness environment, a \emph{tool} consists of a callable function along with associated metadata and a schema specifying its input and output types. When invoked, the function either returns a value of the schema-specified output type or returns a failure message. We define \emph{tool groups} as sets of functionally equivalent tools with the same schema and backend operation but different names and descriptions. A \emph{skill} is a procedure paired with one tool group, and the agent learns when and how to invoke tools through a skill's instructions. We mount these as \emph{redundant tool-skill sets} (Figure~\ref{fig:ecology}), so that entries for the same task family differ only in description and hidden reliability.

\paragraph{Cue neutrality.} For our pure WCST-like experiments, we use neutral textual descriptions for all skills, tools, and tasks. User simulations and templates use task-relevant and neutral vocabulary while varying subject and setting details. Following WCST, there is no lexical cue toward any specific tool group, and agents discover the active tool group by learning from binary tool feedback. We vary the surface wording of descriptions to keep agents' choice of active tool group from simply tracking the wording of the template. Appendix~\ref{app:neutrality} reports an audit for signs of evaluation awareness that may result from modern LLM post-training.

\paragraph{Interventions against neutrality.} Furthermore, we perform multiple interventions that depart from the pure set-shifting paradigm. In the first setup, we vary the cueing with \emph{set framing}, altering skill descriptions to cast the groups relative to one another as either competing providers or a complementary stack. In an alternate setup, we apply a more instructional \emph{policy prompt} to see if the agent adapts accordingly. We detail both setups in \S\ref{sec:results-linear} and \S\ref{sec:results-persona}.
\subsection{Domains and Dataset Construction}
\label{sec:dataset}

We instantiate our test in three domains: scheduling and coordination (3 groups $\times$ 5 tools), DevOps incident triage (4 groups $\times$ 5 tools), and multi-cloud storage (5 groups $\times$ 2 tools). Tools in our tests are simulated, with side-effect actions implemented as mocks, and each mounted set instantiates every relation from Table~\ref{tab:conflict-types}. Scheduling maintains neutral cueing throughout. We build both framings over the same toolset to form the matched pairs used in \S\ref{sec:results-linear} for the DevOps and multi-cloud domains.

Each domain mounts one skill per tool group, with the skills loaded simultaneously. Names within a group draw from a single theme and the naming is parallel across groups. Each domain also includes a support tool group and side-effect tools, which the verifier blocks unless the user explicitly asks for that action. Appendix~\ref{app:construction} documents the framework construction in more detail and Appendix~\ref{app:personas} provides an example of a generated tool-skill set.

\section{Study Design and Evaluation}

\subsection{Harness and Environment}

\paragraph{Setup.} We use Hermes Agent (v0.16.0) \citep{nousresearch2026hermes}, an open-source agentic harness with an extensive skill and tool mechanism. The study focuses on four models of varying sizes and capacity. We use two open-weight models, \texttt{mimo-v2.5} \citep{xiaomi2026mimo} and \texttt{deepseek-v4-pro} \citep{deepseek2026v4} in combination with three proprietary models: \texttt{gemini-3.5-flash-lite} \citep{google2026geminiflashlite}, a fast model for agentic use, and the frontier model \texttt{gpt-5.5} \citep{openai2026gpt55}. We serve the open-weight models and \texttt{gemini-3.5-flash-lite} through OpenRouter and \texttt{gpt-5.5} through the OpenAI API. Every agent uses the default Hermes system prompt (as of v0.16.0). Appendix~\ref{app:setup} gives the sampling settings and remaining configuration.

\subsection{Study Design}
\begin{figure}[t]
\centering
\begin{minipage}[t]{0.48\textwidth}
\centering
\resizebox{0.75\linewidth}{!}{%
\begin{tikzpicture}[
  scale=0.75, transform shape,
  font=\small,
  segnode/.style   ={rectangle, draw, thick, rounded corners=1mm,
                     minimum width=12mm, minimum height=7.5mm,
                     align=center, font=\footnotesize, inner sep=2pt},
  noflipnode/.style={segnode, dashed, fill=gray!12},
  flipedge/.style  ={->, >=Stealth, semithick},
  noflipedge/.style={->, >=Stealth, semithick, gray!60!black, dashed},
  edgelbl/.style   ={midway, font=\tiny, fill=white, inner sep=1pt}
]
  \node[segnode, fill=clusB!25] (root) at (0, 0) {AB\_prefix};
  \node[segnode, fill=clusC!25] (abc) at (3,  1.8) {ABC};
  \node[segnode, fill=clusA!25] (aba) at (3, -1.8) {ABA};
  \node[noflipnode]             (abb) at (3, -3.6) {ABB};
  \node[font=\tiny, gray!55!black, anchor=north] at (3,-4.1)
        {no-shift control};
  \node[segnode, fill=clusA!25] (abca) at (6,  3.0) {ABCA};
  \node[segnode, fill=clusB!25] (abcb) at (6,  1.8) {ABCB};
  \node[noflipnode]             (abcc) at (6,  0.6) {ABCC};
  \node[segnode, fill=clusC!25] (abac) at (6, -0.6) {ABAC};
  \node[segnode, fill=clusB!25] (abab) at (6, -1.8) {ABAB};
  \node[noflipnode]             (abaa) at (6, -3.0) {ABAA};
  \draw[flipedge]   (root) -- node[edgelbl] {B$\to$C} (abc);
  \draw[flipedge]   (root) -- node[edgelbl] {B$\to$A} (aba);
  \draw[noflipedge] (root) -- node[edgelbl] {B$\to$B} (abb);
  \draw[flipedge]   (abc) -- node[edgelbl] {C$\to$A} (abca);
  \draw[flipedge]   (abc) -- node[edgelbl] {C$\to$B} (abcb);
  \draw[noflipedge] (abc) -- node[edgelbl] {C$\to$C} (abcc);
  \draw[flipedge]   (aba) -- node[edgelbl] {A$\to$C} (abac);
  \draw[flipedge]   (aba) -- node[edgelbl] {A$\to$B} (abab);
  \draw[noflipedge] (aba) -- node[edgelbl] {A$\to$A} (abaa);
  \node[font=\scriptsize, gray!55!black] at (0,4.0) {tier 0};
  \node[font=\scriptsize, gray!55!black] at (3,4.0) {tier 1};
  \node[font=\scriptsize, gray!55!black] at (6,4.0) {tier 2};
  \draw[gray!45, dashed] (-0.8,3.65) -- (6.8,3.65);
\end{tikzpicture}
}
\caption{\textbf{Branched schedule tree.} The shared prefix branches into three tier 1 segments (two shifts plus a no-shift control), and each shift segment branches again into three tier 2 endpoints.}
\label{fig:design-tree}
\end{minipage}\hfill
\begin{minipage}[t]{0.48\textwidth}
\centering
\resizebox{\linewidth}{!}{%
\begin{tikzpicture}[
  font=\small,
  layernode/.style={rectangle, draw=gray!50, fill=gray!10, rounded corners=1.5pt,
                    minimum width=11mm, minimum height=4mm, align=center,
                    font=\scriptsize, inner sep=1.5pt},
  arrowline/.style={->, >=Stealth, semithick, gray!60!black},
  routelbl/.style={anchor=east, font=\footnotesize, align=right, inner sep=1pt},
]
  \node[layernode] (ctx1) at (3.7, 5.40) {context + request};
  \fill[clusA!10, rounded corners=2pt] (1.05, 4.41) rectangle (6.45, 4.99);
  \node[routelbl] at (1.0, 4.70) {$R_*$};
  \node[font=\scriptsize, text=clusA] at (1.55, 4.70) {\faCalendar};
  \node[font=\scriptsize, text=clusA] at (1.83, 4.70) {\faCalendarCheck};
  \node[font=\scriptsize, text=clusA] at (2.11, 4.70) {\faCalendarPlus};
  \node[font=\scriptsize, text=clusA] at (2.39, 4.70) {\faCalendarMinus};
  \node[font=\scriptsize, text=clusA] at (2.67, 4.70) {\faCalendarDay};
  \node[font=\scriptsize, text=gray!35] at (3.15, 4.70) {\faClock};
  \node[font=\scriptsize, text=gray!35] at (3.43, 4.70) {\faHourglassHalf};
  \node[font=\scriptsize, text=gray!35] at (3.71, 4.70) {\faStopwatch};
  \node[font=\scriptsize, text=gray!35] at (3.99, 4.70) {\faClipboardList};
  \node[font=\scriptsize, text=gray!35] at (4.27, 4.70) {\faHistory};
  \node[font=\scriptsize, text=gray!35] at (4.75, 4.70) {\faSearch};
  \node[font=\scriptsize, text=gray!35] at (5.03, 4.70) {\faSearchPlus};
  \node[font=\scriptsize, text=gray!35] at (5.31, 4.70) {\faFilter};
  \node[font=\scriptsize, text=gray!35] at (5.59, 4.70) {\faList};
  \node[font=\scriptsize, text=gray!35] at (5.87, 4.70) {\faRandom};
  \draw[arrowline] (ctx1) -- (3.7, 5.04);

  \node[layernode] (ctx2) at (3.7, 3.80) {context + request};
  \fill[clusA!6, rounded corners=2pt] (1.05, 2.81) rectangle (3.7, 3.39);
  \fill[clusB!6, rounded corners=2pt] (3.7, 2.81) rectangle (6.45, 3.39);
  \node[routelbl] at (1.0, 3.10) {$R_2$};
  \node[font=\scriptsize, text=clusA] at (1.55, 3.10) {\faCalendar};
  \node[font=\scriptsize, text=clusA] at (1.83, 3.10) {\faCalendarCheck};
  \node[font=\scriptsize, text=gray!35] at (2.11, 3.10) {\faCalendarPlus};
  \node[font=\scriptsize, text=gray!35] at (2.39, 3.10) {\faCalendarMinus};
  \node[font=\scriptsize, text=gray!35] at (2.67, 3.10) {\faCalendarDay};
  \node[font=\scriptsize, text=gray!35] at (3.15, 3.10) {\faClock};
  \node[font=\scriptsize, text=gray!35] at (3.43, 3.10) {\faHourglassHalf};
  \node[font=\scriptsize, text=clusB] at (3.71, 3.10) {\faStopwatch};
  \node[font=\scriptsize, text=clusB] at (3.99, 3.10) {\faClipboardList};
  \node[font=\scriptsize, text=clusB] at (4.27, 3.10) {\faHistory};
  \node[font=\scriptsize, text=gray!35] at (4.75, 3.10) {\faSearch};
  \node[font=\scriptsize, text=gray!35] at (5.03, 3.10) {\faSearchPlus};
  \node[font=\scriptsize, text=gray!35] at (5.31, 3.10) {\faFilter};
  \node[font=\scriptsize, text=gray!35] at (5.59, 3.10) {\faList};
  \node[font=\scriptsize, text=gray!35] at (5.87, 3.10) {\faRandom};
  \draw[arrowline] (ctx2) -- (3.7, 3.44);

  \node[layernode] (ctx3) at (3.7, 2.20) {context + request};
  \fill[clusB!10, rounded corners=2pt] (1.05, 1.21) rectangle (6.45, 1.79);
  \node[routelbl] at (1.0, 1.50) {$R_1$};
  \node[font=\scriptsize, text=gray!35] at (1.55, 1.50) {\faCalendar};
  \node[font=\scriptsize, text=gray!35] at (1.83, 1.50) {\faCalendarCheck};
  \node[font=\scriptsize, text=gray!35] at (2.11, 1.50) {\faCalendarPlus};
  \node[font=\scriptsize, text=gray!35] at (2.39, 1.50) {\faCalendarMinus};
  \node[font=\scriptsize, text=gray!35] at (2.67, 1.50) {\faCalendarDay};
  \node[font=\scriptsize, text=clusB] at (3.15, 1.50) {\faClock};
  \node[font=\scriptsize, text=clusB] at (3.43, 1.50) {\faHourglassHalf};
  \node[font=\scriptsize, text=clusB] at (3.71, 1.50) {\faStopwatch};
  \node[font=\scriptsize, text=clusB] at (3.99, 1.50) {\faClipboardList};
  \node[font=\scriptsize, text=clusB] at (4.27, 1.50) {\faHistory};
  \node[font=\scriptsize, text=gray!35] at (4.75, 1.50) {\faSearch};
  \node[font=\scriptsize, text=gray!35] at (5.03, 1.50) {\faSearchPlus};
  \node[font=\scriptsize, text=gray!35] at (5.31, 1.50) {\faFilter};
  \node[font=\scriptsize, text=gray!35] at (5.59, 1.50) {\faList};
  \node[font=\scriptsize, text=gray!35] at (5.87, 1.50) {\faRandom};
  \draw[arrowline] (ctx3) -- (3.7, 1.84);

  \draw[gray!45, dashed, thin] (6.5, 1.15) -- (6.5, 5.02);
  \node[font=\tiny, gray!60!black] at (7.04, 5.15) {support};
  \fill[gray!8, rounded corners=2pt] (6.6, 4.41) rectangle (7.5, 4.99);
  \node[font=\scriptsize, text=black!55] at (6.86, 4.70) {\faUserFriends};
  \node[font=\scriptsize, text=black!55] at (7.24, 4.70) {\faMapMarker};
  \fill[gray!8, rounded corners=2pt] (6.6, 2.81) rectangle (7.5, 3.39);
  \node[font=\scriptsize, text=black!55] at (6.86, 3.10) {\faUserFriends};
  \node[font=\scriptsize, text=black!55] at (7.24, 3.10) {\faMapMarker};
  \fill[gray!8, rounded corners=2pt] (6.6, 1.21) rectangle (7.5, 1.79);
  \node[font=\scriptsize, text=black!55] at (6.86, 1.50) {\faUserFriends};
  \node[font=\scriptsize, text=black!55] at (7.24, 1.50) {\faMapMarker};
\end{tikzpicture}
}
\caption{\textbf{Three consecutive turns after an $A \to B$ shift.} Each turn calls a route over the primary groups on the left and the support tools it needs on the right. Classes shown are $R_*$ (no target), $R_2$ (target plus one other), and $R_1$ (target only), with $R_3$ omitted.}
\label{fig:design-routes}
\end{minipage}
\end{figure}

\begin{figure*}[t]
\centering
\begin{tikzpicture}[x=2.7mm, y=1mm, font=\small]
  \fill[clusB!25] (0,9)  rectangle (30,13);
  \fill[clusC!25] (30,9) rectangle (50,13);
  \draw[thick]   (0,9)  rectangle (50,13);
  \node[font=\bfseries, text=clusB!60!black] at (15,11) {B};
  \node[font=\bfseries, text=clusC!60!black] at (40,11) {C};
  \node[font=\scriptsize, anchor=south] at (15,13.2)
        {phase 1: tool group B reliable};
  \node[font=\scriptsize, anchor=south] at (40,13.2)
        {phase 2: tool group C reliable};
  \draw[red!80!black, line width=0.4mm, dashed] (30,-3) -- (30,17.5);
  \node[red!80!black, font=\scriptsize, anchor=south]
        at (30,17.5) {shift at boundary $k$};
  \foreach \t/\c in {%
    1/OK,2/OK,3/OK,4/Mix,5/OK,6/OK,7/OK,8/OK,9/OK,10/Mix,%
    11/OK,12/OK,13/OK,14/OK,15/OK,16/Mix,17/OK,18/OK,19/OK,20/OK,%
    21/OK,22/OK,23/OK,24/OK,25/OK,26/OK,27/OK,28/OK,29/OK,30/OK,%
    31/Old,32/Old,33/Mix,34/Old,35/Mix,36/Mix,37/OK,38/Mix,39/Old,40/Mix,%
    41/OK,42/Mix,43/OK,44/OK,45/Mix,46/OK,47/OK,48/OK,49/OK,50/OK%
  }{\fill[cell\c] ({\t-1},0) rectangle (\t,2);}
  \draw[thick] (0,0) rectangle (50,2);
  \foreach \t in {1,10,20,30,40,50}{
    \draw (\t,0) -- (\t,-1);
    \node[font=\tiny, anchor=north, inner sep=1pt] at (\t,-1) {\t};
  }
  \node[font=\tiny, anchor=west] at (51,-1.5) {turn};
  \draw[decorate, decoration={brace, amplitude=2pt}] (30,2.8) -- (26,2.8);
  \node[font=\tiny, anchor=south] at (28,3.0) {pre-shift plateau};
  \draw[decorate, decoration={brace, amplitude=2pt}] (40,2.8) -- (31,2.8);
  \node[font=\tiny, anchor=south] at (35.5,3.0) {post-shift window};
  \node[font=\tiny, anchor=west] at (-1,-5.5) {Cell:};
  \fill[cellOK]  (3,-6.1)  rectangle (4,-5.0);
  \node[font=\tiny, anchor=west] at (4.3,-5.5) {clean commit};
  \fill[cellMix] (12,-6.1) rectangle (13,-5.0);
  \node[font=\tiny, anchor=west] at (13.3,-5.5) {mixed};
  \fill[cellOld] (19,-6.1) rectangle (20,-5.0);
  \node[font=\tiny, anchor=west] at (20.3,-5.5) {off-target};
\end{tikzpicture}
\caption{\textbf{Behavior per turn across a single boundary.} The top bar shows the reliable tool group across phases, and the strip below shows the route-class chosen each turn.}
\label{fig:design}
\end{figure*}
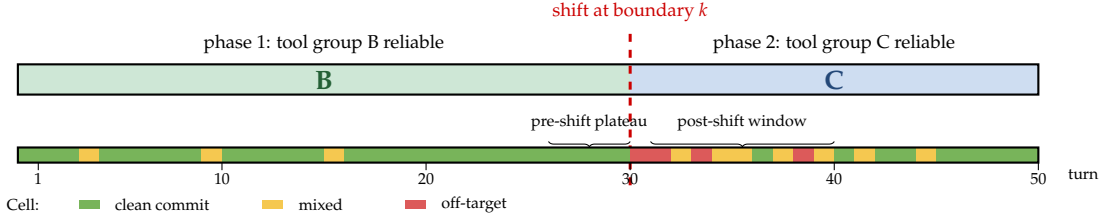

\paragraph{Schedule and tree.}
In our proposed benchmark, the previously reliable tool group becomes unreliable at each shift. The harness context does not provide any warning of the change, and tool descriptions and schemas remain unchanged. Figure~\ref{fig:design} shows the behavior per turn at one such shift boundary. With $M$ tool groups and $K$ branching layers, we represent the schedule as a tree of trajectories (Figure~\ref{fig:design-tree}). Each \emph{trajectory} begins with a shared \emph{prefix}, a common starting point with fixed tool display order and permutation within each group. After a trajectory is complete, the run branches into three trajectories with different settings: a \ehl{clusA!22}{reversal} to a previously used tool group, a \ehl{clusC!22}{novel shift} to a previously unused tool group, and a \ehl{gray!25}{no-shift control} that keeps the previous group reliable.

In this study, we use $M=3$ tool groups and $K=2$ branching layers over 90 turns with 9 endpoints. They contain three boundaries: $b_1$ at turn 15 inside the prefix ($A \to B$), $b_2$ at turn 40 before a tier 1 window, and $b_3$ at turn 65 before a tier 2 window. Each tool group first becomes reliable at a different shift point. Group \eA{} is reliable from turn 1 and \eB{} from turn 16. Group \eC{} first becomes reliable at the tier 1 boundary (turn 41) on novel shift endpoints and is never reliable on no-shift control branches.

We also annotate each target with a \emph{prior use role}, based on the order in which they first appear in the run. $C^A$ is a \ehl{clusA!22}{reversal} target that is reliable early in the prefix and then becomes unreliable. $C^B$ is a \ehl{clusB!22}{reinforced} target that remains reliable through the end of the prefix. $C^C$ is a \ehl{clusC!22}{novel} target that is never reliable in the prefix. Differences in metrics among groups should reflect their first appearances in the run. A role-theme permutation ablation confirms that the differences track prior use rather than the fixed tool group (Appendix~\ref{app:permutation}). 

The tree-based experimental protocol in \S\ref{sec:results-endpoints}, \S\ref{sec:results-depth}, and \S\ref{sec:results-persona} uses the scheduling domain with $n=16$ runs. Every domain also supports a \emph{linear schedule}, a single trajectory with shifts at fixed turns for cost efficiency. We use this schedule for the set framing intervention on DevOps and multi-cloud (\S\ref{sec:results-linear}) and for the replication on a second harness (\S\ref{app:harness}).

\paragraph{Research questions.}
\label{sec:rqs}
In this study, we ask the following:
\begin{enumerate}
\item When the reliable tool group shifts, does the agent commit to the new tool group, and how does each prior-use role affect the resulting pattern?
\item Does the depth of an endpoint, or the path taken to reach it, matter beyond the most recent shift?
\item Does set framing shift which routine the agent settles into?
\item Can an instructional policy prompt affect the lock-in?
\end{enumerate}

\subsection{Evaluation metrics}
\label{sec:eval}

Here we define some metrics over trajectory observations. \emph{Set-shifting accuracy} $\Phi$ measures whether the agent stays on the reliable group across a trajectory. Per-turn \emph{route-class rates} describe the routing profile at each shift. The \emph{task-completion rate} $F$ records whether the reliable work is completed, and the \emph{single-routine rate} measures rigidity. We use these metrics to answer the four research questions we pose.

\paragraph{Set-shifting accuracy.}
For each post-shift window $\mathcal{P}_k$, we define the target share $s_k$ as the fraction of route calls made to the reliable group $C_k$. For a trajectory crossing boundaries $b_1, \ldots, b_K$,
\begin{equation}
\Phi \;=\; \prod_{k=1}^{K} s_k.
\end{equation}
The product is high only when the agent routes reliably after every shift. Equivalently, suppose we sample route calls from each post-shift window, then $\Phi$ is the probability that every sampled call belongs to its target group. Because $\Phi$ is cumulative, we compare models at the same endpoint and use the single-window share $s_k$ to measure an individual boundary.

\paragraph{Route classes.}
We classify each turn by the set of tool groups its calls touch (Figure~\ref{fig:design-routes}). $R_1$ denotes \emph{target only}, $R_2$ \emph{target plus one other group}, $R_3$ \emph{target plus both other groups}, and $R_*$ \emph{no target}. With one tool call per included group, these classes correspond to $s \in \{1, 0.5, 0.33, 0\}$. We call $R_2$ and $R_3$ \emph{breadth routes} because they retain the target while adding one or both other groups. $R_*$ denotes \emph{off-target routing}, which omits the currently reliable group. The $R$-classes are turn-level labels, whereas $s_k$ is their window-level average under this mapping. Because $s_k$ weights every route call equally, turns with many calls receive more weight. The model ordering is unchanged under a turn-normalized share (Appendix~\ref{app:callweight}).

\paragraph{Task completion.}
We calculate the proportion of turns that the verifier marks as task-completing, $F$. A turn counts when the agent calls at least one tool from the reliable group, uses every required support tool group, and invokes no side-effect tools. A turn that mixes the reliable group with unreliable groups still counts as complete.

\paragraph{Single-routine rate.}
A trajectory holds a \emph{single routine} when it calls the same set of tool groups on every route-bearing turn of a post-shift window, and the single-routine rate is the fraction of windows for which this holds. We use this metric to measure rigidity.

\paragraph{Uncertainty.} All reported $\Phi$ values carry bootstrap 95\% intervals, with paired permutation tests for sibling contrasts (Appendix~\ref{app:setup}).

\section{Experimental Results and Analysis}
\label{sec:results}

\subsection{Target history interacts with model-specific commitment (RQ1)}
\label{sec:results-endpoints}

\paragraph{Results.} An agent's response to a shift depends on the target's prior use, but this dependence differs by model. By mean set-shifting accuracy, \texttt{mimo} orders the targets \ehl{clusA!22}{$C^A$} $>$ \ehl{clusB!22}{$C^B$} $>$ \ehl{clusC!22}{$C^C$} and has low variance. Its accuracy is lowest at the never-seen target. \texttt{deepseek-v4-pro} instead orders them \ehl{clusB!22}{$C^B$} $>$ \ehl{clusA!22}{$C^A$} $\approx$ \ehl{clusC!22}{$C^C$} and has high variance at every target (Table~\ref{tab:fidelity-ranking}).

\begin{table*}[t]
\centering
\footnotesize
\setlength{\tabcolsep}{3pt}
\resizebox{\textwidth}{!}{%
\begin{tabular}{l|cccc|cccc|cccc|cccc}
\toprule
& \multicolumn{4}{c|}{\texttt{mimo-v2.5}} & \multicolumn{4}{c|}{\texttt{deepseek-v4-pro}} & \multicolumn{4}{c|}{\texttt{gemini-3.5-flash-lite}} & \multicolumn{4}{c}{\texttt{gpt-5.5}} \\
\cmidrule(lr){2-5}\cmidrule(lr){6-9}\cmidrule(lr){10-13}\cmidrule(lr){14-17}
\textbf{Endpoint} & $\Phi|s_k$ & $R_*$ & $R_3$ & $F$ & $\Phi|s_k$ & $R_*$ & $R_3$ & $F$ & $\Phi|s_k$ & $R_*$ & $R_3$ & $F$ & $\Phi|s_k$ & $R_*$ & $R_3$ & $F$ \\
\midrule
\multicolumn{17}{l}{\textit{Tier 1} ($s_k$ at $b_2$)} \\
\eA\eB\eA & $.31{\pm}.06|.82$ & .00 & .03 & .99 & $.41{\pm}.16|.53$ & .25 & .03 & .75 & $.45{\pm}.14|.74$ & .17 & .00 & .74 & $.27{\pm}.05|.52$ & .00 & .13 & 1.00 \\
\eA\eB\eB$^\dagger$ & $.22{\pm}.14|.40$ & .00 & .18 & 1.00 & $.82{\pm}.16|.86$ & .06 & .00 & .94 & $.64{\pm}.21|.70$ & .19 & .00 & .80 & $.28{\pm}.06|.52$ & .02 & .16 & .98 \\
\eA\eB\eC & $.06{\pm}.04|.17$ & .39 & .46 & .60 & $.38{\pm}.18|.40$ & .34 & .20 & .66 & $.15{\pm}.12|.23$ & .57 & .15 & .39 & $.14{\pm}.03|.25$ & .13 & .51 & .87 \\
\midrule
\multicolumn{17}{l}{\textit{Tier 2} ($s_k$ at $b_3$)} \\
\eA\eB\eA\eA$^\dagger$ & $.25{\pm}.05|.83$ & .00 & .03 & 1.00 & $.40{\pm}.16|.68$ & .19 & .02 & .75 & $.44{\pm}.15|.99$ & .01 & .00 & .81 & $.16{\pm}.05|.55$ & .00 & .11 & 1.00 \\
\eA\eB\eA\eB & $.15{\pm}.07|.40$ & .00 & .20 & .99 & $.34{\pm}.14|.72$ & .06 & .02 & .88 & $.19{\pm}.10|.35$ & .42 & .00 & .58 & $.13{\pm}.03|.45$ & .08 & .14 & .92 \\
\eA\eB\eA\eC & $.05{\pm}.02|.20$ & .31 & .52 & .68 & $.17{\pm}.12|.27$ & .50 & .07 & .44 & $.11{\pm}.11|.15$ & .77 & .02 & .23 & $.06{\pm}.02|.21$ & .31 & .38 & .69 \\
\eA\eB\eC\eA & $.04{\pm}.02|.80$ & .00 & .10 & .99 & $.24{\pm}.13|.52$ & .28 & .06 & .66 & $.07{\pm}.07|.64$ & .29 & .01 & .70 & $.06{\pm}.01|.46$ & .01 & .32 & .99 \\
\eA\eB\eC\eB & $.03{\pm}.02|.39$ & .00 & .28 & .99 & $.30{\pm}.16|.66$ & .13 & .08 & .80 & $.07{\pm}.07|.41$ & .39 & .00 & .61 & $.07{\pm}.02|.51$ & .04 & .20 & .96 \\
\eA\eB\eC\eC$^\dagger$ & $.02{\pm}.01|.20$ & .31 & .50 & .68 & $.37{\pm}.18|.60$ & .25 & .09 & .68 & $.13{\pm}.12|.37$ & .50 & .12 & .49 & $.06{\pm}.02|.39$ & .10 & .29 & .90 \\
\midrule
\emph{single-routine rate} & \multicolumn{4}{c|}{66\%} & \multicolumn{4}{c|}{50\%} & \multicolumn{4}{c|}{66\%} & \multicolumn{4}{c}{9\%} \\
\bottomrule
\end{tabular}%
}
\caption{\textbf{Per-model behavior on the scheduling tree.} Route-class rates ($R_*$, $R_3$), completion $F$, final-window target share $s_k$, and set-shifting accuracy $\Phi$ across the four primary models ($n=16$ each). The $\Phi|s_k$ column pairs $\Phi$ (mean $\pm$ bootstrap 95\% CI half-width) with the final-window share $s_k$, and the endpoint's final letter gives the target group. $\dagger$ marks a no-shift control. Definitions are in \S\ref{sec:eval}.}
\label{tab:fidelity-ranking}
\end{table*}

\begin{figure*}[t]
\centering
\includegraphics[width=0.78\textwidth]{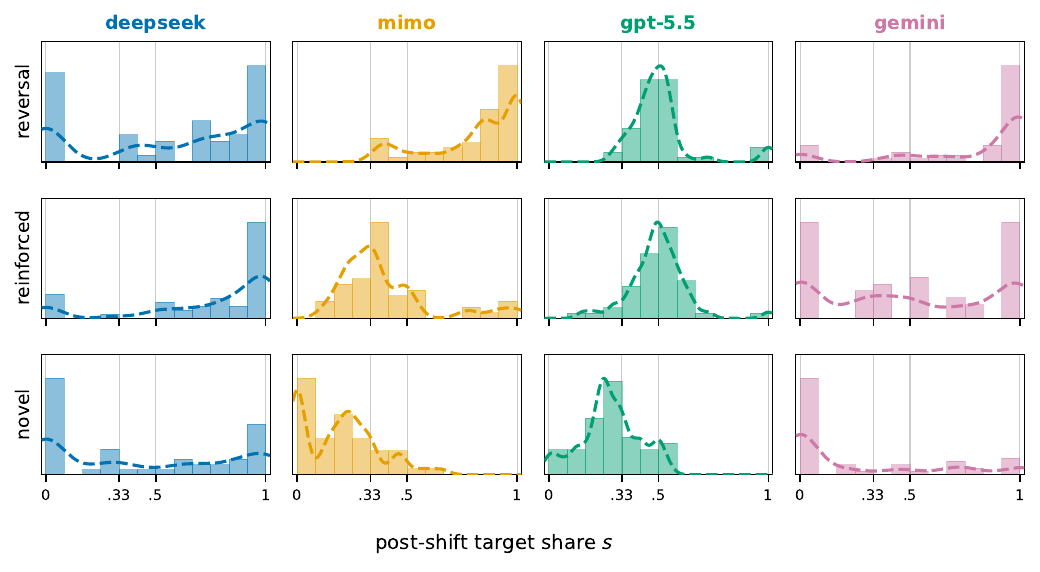}
\caption{\textbf{Post-shift target share splits by prior-use role.} Histograms pooled over the three endpoints in each role. Columns are the models and rows are the \ehl{clusA!22}{reversal}, \ehl{clusB!22}{reinforced}, and \ehl{clusC!22}{novel} roles, with vertical guides at $\{0, \tfrac{1}{3}, \tfrac{1}{2}, 1\}$. \texttt{deepseek-v4-pro} splits to the extremes, \texttt{mimo-v2.5} concentrates on the middle, \texttt{gpt-5.5} holds a single mode, and \texttt{gemini-3.5-flash-lite} is bimodal with values near $0$ on novel targets.}
\label{fig:s-target-density}
\end{figure*}

\paragraph{Post-shift target shares concentrate on a few discrete values.}
After a shift, target share tends to approach $s \in \{0, \frac{1}{3}, \frac{1}{2}, 1\}$. The resulting value depends on both the model and the target's prior-use role (Figure~\ref{fig:s-target-density}). Appendix~\ref{app:density} gives the complete endpoint-level results, and Appendix~\ref{app:reasoning} gives example trajectories.

\paragraph{Models differ in how rigidly they maintain these routines.}
\texttt{mimo}, \texttt{deepseek-v4-pro}, and \texttt{gemini-3.5-flash-lite} follow a single routine much more often than \texttt{gpt-5.5} (Table~\ref{tab:fidelity-ranking}). For example, \texttt{gemini} often uses the target-only routine after reversal shifts but an off-target routine after novel shifts. Each recurring value corresponds to a stable routine \citep{botvinick2004schema}.

\paragraph{Models maintain different routines.}
The two open-weight models maintain routines at different levels, explaining their contrasting variance. \texttt{mimo}'s low variance reflects a routine determined by the target's prior use. This pattern represents partial commitment. By contrast, \texttt{deepseek-v4-pro}'s high variance reflects prefix-level commitment to whichever group it selects early. A prefix that selects the reliable group retains it to the end ($s \approx 1$), whereas a prefix that does not remains off-target ($s \approx 0$). This split persists across the tier boundary and prior use barely changes the outcome (Figure~\ref{fig:phi-over-time}).

On the other hand, \texttt{gpt-5.5} follows a single routine in only $9\%$ of windows yet maintains high task completion, including $F=0.87$ at the novel target. Its target share remains low because it calls other groups alongside the target. Completion falls to $0.69$ after a reversal (ABAC).

These set of behaviors define two routing profiles that repeats across our results. In the \emph{concentration} profile, seen on \texttt{deepseek}'s low-share prefixes, a model commits to one group and routes off-target ($R_*$) when that group becomes unreliable. In the \emph{retention} profile, seen with \texttt{gpt-5.5}, a model keeps the reliable group while adding breadth ($R_2$ and $R_3$). We find that the distinction persists under an independent harness (Appendix~\ref{app:harness}).

\subsection{Matched scheduled paths add little beyond the current shift (RQ2)}
\label{sec:results-depth}

Endpoint depth adds little beyond the current shift. Because $\Phi$ multiplies one share per window, it decreases with depth by construction. The final window share $s_k$ remains stable. For example, \texttt{mimo} reaches $s_k=0.82$ at \eA\eB\eA and $s_k=0.83$ one level deeper at \eA\eB\eA\eA.

The preceding path also has little effect. The paths \eA$\to$\eB$\to$\eA$\to$\eB and \eA$\to$\eB$\to$\eC$\to$\eB reach the same reinforced target through different intermediate groups. Across the tested models, their final shares differ by at most $0.06$.

We find that the current shift has a larger effect: for \texttt{deepseek}, shifting away from the reinforced target reduces $\Phi$ by approximately $0.40$ relative to the matched no-shift control.\footnote{Paired-permutation $\Delta\Phi$ tests give $p = 6\times10^{-4}$ for the reversal (\eA\eB\eB{} vs \eA\eB\eA) and $6\times10^{-5}$ for the novel shift (\eA\eB\eB{} vs \eA\eB\eC).} Behavior therefore depends more on the current shift than on endpoint depth or the path.

\subsection{Set framing shifts routine occupancy (RQ3)}
\label{sec:results-linear}

\paragraph{Setup.}
We vary set framing in the DevOps ($M=4$) and multi-cloud ($M=5$) domains, using 15 turns per phase under the linear schedule. Each domain uses a matched pair over the same toolset that differs only in framing (Appendix~\ref{app:framing}). The complementary condition presents the groups as parts of one service, each returning its own fields. The competing condition presents them as rival providers whose reports share one identifier. We measure \emph{routine width} $\bar{w}$, the mean number of distinct tool groups called per turn.

\paragraph{Results.}
Competing framing narrows the routine for every model in both domains (Table~\ref{tab:paired-framing}). Complementary framing invites a mixed routine spanning all $M$ groups, whereas competing framing invites commitment to one. In 53 of the 64 matched pairs, the competing side is narrower (sign test $p = 1.0 \times 10^{-7}$). The effect is sharpest for \texttt{gpt-5.5}, whose multi-cloud routine contracts from all five groups to about one. It also appears for \texttt{gemini-3.5-flash-lite} on multi-cloud ($3.5 \to 1.2$).

Narrowing the routine does not always improve set-shifting. \texttt{deepseek-v4-pro}, \texttt{mimo-v2.5}, and \texttt{gpt-5.5} commit toward the reliable group. Competing framing therefore raises both their target share and set-shifting accuracy. For example, \texttt{deepseek}'s DevOps $\Phi$ rises from $0.02$ to $0.26$. \texttt{gemini} also narrows sharply on multi-cloud, but it selects a non-target group and its $\Phi$ remains near zero. Framing therefore changes \emph{commitment}, which improves accuracy only when the model commits to the target.

\begin{table}[t]
\centering
\begin{minipage}[t]{0.56\textwidth}
\centering
\footnotesize
\setlength{\tabcolsep}{3.5pt}
\begin{tabular}{@{}ll cc cc@{}}
\toprule
& & \multicolumn{2}{c}{width $\bar{w}$} & \multicolumn{2}{c}{$\Phi$} \\
\cmidrule(lr){3-4} \cmidrule(lr){5-6}
\textbf{Model} & \textbf{Domain} ($M$) & cpl. & cpt. & cpl. & cpt. \\
\midrule
\texttt{deepseek} & DevOps (4)      & 3.49 & \textbf{1.98} & 0.02 & 0.26 \\
                  & Multi-cloud (5) & 4.32 & \textbf{2.90} & 0.01 & 0.18 \\
\midrule
\texttt{mimo}     & DevOps (4)      & 3.81 & \textbf{2.80} & 0.03 & 0.11 \\
                  & Multi-cloud (5) & 4.33 & \textbf{2.71} & 0.01 & 0.03 \\
\midrule
\texttt{gpt-5.5}  & DevOps (4)      & 4.00 & \textbf{2.06} & 0.00 & 0.10 \\
                  & Multi-cloud (5) & 5.00 & \textbf{1.15} & 0.00 & 0.28 \\
\midrule
\texttt{gemini}   & DevOps (4)      & 2.82 & \textbf{2.57} & 0.01 & 0.01 \\
                  & Multi-cloud (5) & 3.47 & \textbf{1.23} & 0.00 & 0.04 \\
\bottomrule
\end{tabular}
\caption{\textbf{Competing framing narrows the routine.} Matched framing pairs on the DevOps and multi-cloud domains, $n=8$ per cell. Width $\bar{w}$ is the mean number of distinct tool groups per turn, at most $M$. cpl./cpt.\ mark the complementary/competing framings.}
\label{tab:paired-framing}
\end{minipage}\hfill
\begin{minipage}[t]{0.42\textwidth}
\centering
\footnotesize
\setlength{\tabcolsep}{4.5pt}
\begin{tabular}{@{}lcccc@{}}
\toprule
\textbf{Model} & \textbf{II} & \textbf{reg.}\,$\Phi$ & \textbf{adapt.}\,$\Phi$ & \textbf{$R_*$ nov.} \\
\midrule
\texttt{gpt-5.5}  & 56 & .14 & \textbf{.86} & $.18\!\to\!.00$ \\
\texttt{deepseek} & 44 & .38 & \textbf{.74} & $.36\!\to\!.00$ \\
\texttt{mimo}     & 37 & .13 & .15          & $.34\!\to\!.44$ \\
\texttt{gemini}   & 36 & .25 & .32          & $.61\!\to\!.52$ \\
\bottomrule
\end{tabular}
\caption{\textbf{Prompt response differs across the tested capability range.} $\Phi$ is the mean set-shifting accuracy under the regular assistant ($n{=}16$) and adaptive prompt ($n{=}8$), $R_*$ the novel-window off-target rate, and \textbf{II} the Artificial Analysis Intelligence Index.}
\label{tab:rq4-gating}
\end{minipage}
\end{table}

\subsection{An adaptive prompt mitigates the lock-in (RQ4)}
\label{sec:results-persona}

\paragraph{Setup.}

We append a trajectory-local policy prompt that instructs the agent to reuse the most recently successful tool and to switch groups after a failure (Appendix~\ref{app:personas}). We evaluate the prompt across the panel. We then compare prompt response with the Artificial Analysis Intelligence Index, which we use as a coarse capability proxy \citep{artificialanalysis2026}.

\paragraph{Results.}

The adaptive prompt substantially improves set-shifting for \texttt{gpt-5.5} and \texttt{deepseek-v4-pro}, but has little effect on \texttt{mimo-v2.5} and \texttt{gemini-3.5-flash-lite} (Table~\ref{tab:rq4-gating}). Mean accuracy rises from $0.14$ to $0.86$ for \texttt{gpt-5.5} and from $0.38$ to $0.74$ for \texttt{deepseek-v4-pro}. Novel-window off-target routing falls to zero for both. On \texttt{deepseek}, the prompt also overrides the bimodality of \S\ref{sec:results-endpoints}. The regular assistant leaves 9 of 16 ABAC trajectories at $s \leq 0.2$, whereas no adaptive trajectory remains in that low mode. 

\begin{wrapfigure}{r}{0.52\textwidth}
\centering
\includegraphics[width=\linewidth]{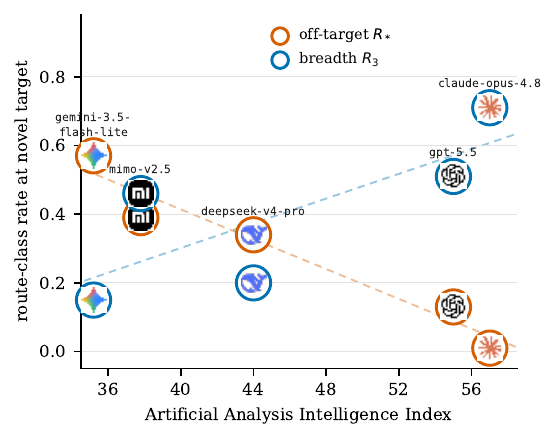}
\caption{\textbf{Higher capability shifts the routing profile from concentration toward retention.} At novel targets, a higher Artificial Analysis Intelligence Index coincides with less off-target routing ($R_*$) and more breadth ($R_3$).}
\label{fig:intel}
\end{wrapfigure}

The same prompt barely changes set-shifting accuracy for the other two models and off-target routing persists. On \texttt{mimo}, the prompt increases novel-window $R_*$. Within our panel, model capability coincides both with the observed routing profile and with prompt response (Figure~\ref{fig:intel}). These models may require stronger prompting, external control, or a training-based intervention. We leave those alternatives to future work.

\section{Related Work}

\paragraph{Tool-use benchmarks and redundant tools.}
Single-session tool-use benchmarks evaluate tool selection and execution \citep{yao2024taubench,barres2025tau2,huang2024metatool,patil2023gorilla,patil2025bfcl,qin2023toolllm,liu2024agentbench}. Separate bias studies report fixation \citep{blankenstein2026biasbusters,wei2024selection}, and \citet{faghih2025toolprefs} show that for two tools with identical arguments, description edits alone shift selection. Concurrent work explore adjacent axes: \citet{lin2025toolscope} merge and filter redundant tool inventories, \citet{liu2025costbench} evaluate replanning under dynamic tool failures and costs, and \citet{chen2026inertia} study failure to update multi-turn reasoning and correct it with single-turn-anchored reinforcement learning. We differ by combining \emph{within-session} shifts of group reliability with resumed, branched session state on top of an agentic harness \citep{yao2023react,schick2023toolformer,shinn2023reflexion,park2023generative,wang2023voyager,packer2023memgpt}. The reinforcement-learning remedy of \citet{chen2026inertia} is a candidate training-based fix for the models that resist our prompt, such as \texttt{mimo}.
\paragraph{Path dependence and policy effects.}
\citet{lee2026capable} documents
 path dependence within a trajectory, where one off-route call raises the next call's off-route probability. We extend this to runs with multiple shifts that separate perseveration from adaptation to the new reliable tool group. The exploration deficits documented by \citet{krishnamurthy2024exploration} predict the failure we observe on \texttt{mimo}. \citet{cao2026persona} document agentic degradation under persona/policy cues, but do not measure perseveration over time.

\paragraph{Choice perseveration and continual learning.}
The WCST decomposes set-shifting failure into perseverative errors and failure-to-maintain set \citep{heaton1993wcst}, formalized as a framework called Value-Plus-Perseveration \citep{worthy2013perseveration}. Functional fixedness is the spatial analog of Einstellung's procedural rigidity \citep{duncker1945functional}. The same construct surfaces in continual learning as primacy bias \citep{nikishin2022primacy} and loss of plasticity \citep{dohare2024plasticity}. In this work, we import the change protocol into an LLM agent harness.

\paragraph{Cognitive tests on LLMs.}
The closest cognitive tests for LLMs \citep{li2024reflectionbench,hao2025vlmwcst,li2025nearoptimal} port WCST and set-shifting probes into multimodal settings, but these works evaluate raw model behavior without context. We close this gap by testing the agentic version, where the ``cards'' are functionally callable tools and the branched tree of resumed sessions carries the harness state these works omit.

\clearpage
\section{Discussion}
\label{sec:discussion}

\paragraph{What does a low set-shifting accuracy actually mean?}
A low set-shifting accuracy does not always mean the same failure. In the retention profile, the reliable group is still called along with alternatives ($R_2$ and $R_3$), so the task can still complete despite a lower target share. For \texttt{gpt-5.5}, for example, $\Phi$ is among the lowest overall, yet its departures are primarily target-retaining rather than target-omitting, so its completion rate $F$ is among the best. $\Phi$ measures how much of an agent's attention lands on the reliable group, while the route-class rates and $F$ show how it departs and whether it still completes the task.

\paragraph{Which routing profile is right for a deployed agent?}
In actual agent use, target omission may introduce completion risk from backend drift, with agents retaining their previous routes when they are no longer appropriate \citep{wang2025deprecated}. While target-retaining breadth preserves coverage, it increases the call footprint, potentially causing higher latency and rate limit pressure \citep{liu2025costbench}. When the reliability of the results is not guaranteed, redundant calls provide fault tolerance \citep{avizienis1985nversion,dean2013tail}. The preferred profile then depends on an agent application's reliability and latency. Evaluations should report completion, target retention, and route breadth separately, and maximal focus on the reliable group is not always desirable.

\paragraph{Can a prompt intervention alter both profiles?}
Set framing and the policy prompt help the two capable models but barely move the weaker two. A durable fix likely has to act at training time rather than through prompting. Multi-turn tool-use RL \citep{singh2025artist} is a natural candidate for the models that resist our prompt, and continual learning remedies for primacy bias and loss of plasticity \citep{nikishin2022primacy,dohare2024plasticity} target the same lock-in. We leave these training-based methods to future work.

\paragraph{What does robust adaptation require?}
An off-target return to the group that was reliable just before the shift counts as perseveration only when the agent had actually settled on that group first. Other off-target trajectories may instead mean the agent never settled on a single group, or is still exploring older ones \citep{cohen2007explore}. In retentive routes, the agent found the working route but keeps calling alternatives alongside it from lingering uncertainty and hedging. Robust adaptation entails shifting to the newly reliable route and inhibiting the alternatives once the evidence turns against them \citep{miyake2000unity,diamond2013executive}.

\section{Conclusion}

We introduce a longitudinal behavioral test of set-shifting in which the reliable tool group changes silently with a matched no-shift control at every shift. Across the panel, agents adopt a small number of recurring routines and separate into multiple routing profiles. Future work could explore the potential of making tool reliability stochastic so that no single outcome marks a shift, and study how memory affects routing over longer trajectories for more realistic agent use.

\bibliographystyle{colm2026_conference}
\bibliography{custom}

\begin{thebibliography}{48}
\providecommand{\natexlab}[1]{#1}
\providecommand{\url}[1]{\texttt{#1}}
\expandafter\ifx\csname urlstyle\endcsname\relax
  \providecommand{\doi}[1]{doi: #1}\else
  \providecommand{\doi}{doi: \begingroup \urlstyle{rm}\Url}\fi

\bibitem[{Artificial Analysis}(2026)]{artificialanalysis2026}
{Artificial Analysis}.
\newblock Artificial analysis intelligence index.
\newblock \url{https://artificialanalysis.ai}, 2026.
\newblock Aggregate capability index over public benchmarks. Accessed August
  2026.

\bibitem[Avizienis(1985)]{avizienis1985nversion}
Algirdas Avizienis.
\newblock The {N}-version approach to fault-tolerant software.
\newblock \emph{IEEE Transactions on Software Engineering}, SE-11\penalty0
  (12):\penalty0 1491--1501, 1985.

\bibitem[Barres et~al.(2025)Barres, Dong, Ray, Si, and
  Narasimhan]{barres2025tau2}
Victor Barres, Honghua Dong, Soham Ray, Xujie Si, and Karthik Narasimhan.
\newblock $\tau^2$-bench: Evaluating conversational agents in a dual-control
  environment, 2025.

\bibitem[Blankenstein et~al.(2026)Blankenstein, Yu, Li, Plachouras, Sengupta,
  Torr, Gal, Paren, and Bibi]{blankenstein2026biasbusters}
Thierry Blankenstein, Jialin Yu, Zixuan Li, Vassilis Plachouras, Sunando
  Sengupta, Philip Torr, Yarin Gal, Alasdair Paren, and Adel Bibi.
\newblock {BiasBusters}: Uncovering and mitigating tool selection bias in large
  language models.
\newblock In \emph{International Conference on Learning Representations}, 2026.

\bibitem[Botvinick \& Plaut(2004)Botvinick and Plaut]{botvinick2004schema}
Matthew Botvinick and David~C. Plaut.
\newblock Doing without schema hierarchies: A recurrent connectionist approach
  to normal and impaired routine sequential action.
\newblock \emph{Psychological Review}, 111\penalty0 (2):\penalty0 395--429,
  2004.

\bibitem[Cao et~al.(2026)Cao, Sun, and Yue]{cao2026persona}
Linbo Cao, Lihao Sun, and Yang Yue.
\newblock From biased chatbots to biased agents: Examining role assignment
  effects on {LLM} agent robustness, 2026.

\bibitem[Chen et~al.(2026)Chen, Zhang, Guo, and Zou]{chen2026inertia}
Xingwu Chen, Zhanqiu Zhang, Steven~Y. Guo, and Difan Zou.
\newblock Breaking contextual inertia: Reinforcement learning with single-turn
  anchors for stable multi-turn interaction.
\newblock In \emph{Findings of the Association for Computational Linguistics:
  ACL 2026}, pp.\  6302--6318, San Diego, California, United States, July 2026.
  Association for Computational Linguistics.
\newblock \doi{10.18653/v1/2026.findings-acl.313}.
\newblock URL \url{https://aclanthology.org/2026.findings-acl.313/}.

\bibitem[Cohen et~al.(2007)Cohen, McClure, and Yu]{cohen2007explore}
Jonathan~D. Cohen, Samuel~M. McClure, and Angela~J. Yu.
\newblock Should {I} stay or should {I} go? how the human brain manages the
  trade-off between exploitation and exploration.
\newblock \emph{Philosophical Transactions of the Royal Society B},
  362\penalty0 (1481):\penalty0 933--942, 2007.

\bibitem[Dean \& Barroso(2013)Dean and Barroso]{dean2013tail}
Jeffrey Dean and Luiz~Andr\'{e} Barroso.
\newblock The tail at scale.
\newblock \emph{Communications of the ACM}, 56\penalty0 (2):\penalty0 74--80,
  2013.

\bibitem[{DeepSeek-AI}(2026)]{deepseek2026v4}
{DeepSeek-AI}.
\newblock {DeepSeek-V4}: Towards highly efficient million-token context
  intelligence, 2026.
\newblock Model card at
  \url{https://huggingface.co/deepseek-ai/DeepSeek-V4-Pro}.

\bibitem[Diamond(2013)]{diamond2013executive}
Adele Diamond.
\newblock Executive functions.
\newblock \emph{Annual Review of Psychology}, 64:\penalty0 135--168, 2013.

\bibitem[Dohare et~al.(2024)Dohare, Hernandez-Garcia, Lan, Rahman, Mahmood, and
  Sutton]{dohare2024plasticity}
Shibhansh Dohare, J.~Fernando Hernandez-Garcia, Qingfeng Lan, Parash Rahman,
  A.~Rupam Mahmood, and Richard~S. Sutton.
\newblock Loss of plasticity in deep continual learning.
\newblock \emph{Nature}, 632:\penalty0 768--774, 2024.

\bibitem[Duncker(1945)]{duncker1945functional}
Karl Duncker.
\newblock On problem-solving.
\newblock \emph{Psychological Monographs}, 58\penalty0 (5):\penalty0 i--113,
  1945.

\bibitem[{Earendil Works}(2026)]{pi2026agent}
{Earendil Works}.
\newblock pi: An open-source {AI} agent toolkit.
\newblock \url{https://github.com/earendil-works/pi}, 2026.
\newblock Coding agent with its own tool-calling loop and context manager, MIT
  license. Accessed August 2026.

\bibitem[Faghih et~al.(2025)Faghih, Wang, Cheng, Bharti, Sriramanan,
  Balasubramanian, Hosseini, and Feizi]{faghih2025toolprefs}
Kazem Faghih, Wenxiao Wang, Yize Cheng, Siddhant Bharti, Gaurang Sriramanan,
  Sriram Balasubramanian, Parsa Hosseini, and Soheil Feizi.
\newblock Tool preferences in agentic {LLMs} are unreliable.
\newblock In \emph{Proceedings of the 2025 Conference on Empirical Methods in
  Natural Language Processing}, 2025.

\bibitem[{Google DeepMind}(2026)]{google2026geminiflashlite}
{Google DeepMind}.
\newblock {Gemini 3.5 Flash-Lite} model card.
\newblock
  \url{https://deepmind.google/models/model-cards/gemini-3-5-flash-lite/},
  2026.
\newblock accessed August 2026.

\bibitem[Grant \& Berg(1948)Grant and Berg]{grant1948wcst}
David~A. Grant and Esta~A. Berg.
\newblock A behavioral analysis of degree of reinforcement and ease of shifting
  to new responses in a {Weigl}-type card-sorting problem.
\newblock \emph{Journal of Experimental Psychology}, 38\penalty0 (4):\penalty0
  404--411, 1948.

\bibitem[Hao et~al.(2025)Hao, Alexandre, and Yu]{hao2025vlmwcst}
Guangfu Hao, Frederic Alexandre, and Shan Yu.
\newblock Visual large language models exhibit human-level cognitive
  flexibility in the {Wisconsin Card Sorting Test}, 2025.

\bibitem[Heaton et~al.(1993)Heaton, Chelune, Talley, Kay, and
  Curtiss]{heaton1993wcst}
Robert~K. Heaton, Gordon~J. Chelune, Jack~L. Talley, Gary~G. Kay, and Glenn
  Curtiss.
\newblock \emph{Wisconsin Card Sorting Test Manual: Revised and Expanded}.
\newblock Psychological Assessment Resources, 1993.

\bibitem[Huang et~al.(2024)Huang, Shi, Li, Fan, Wu, Zhang, Liu, Zhou, Wan,
  Gong, and Sun]{huang2024metatool}
Yue Huang, Jiawen Shi, Yuan Li, Chenrui Fan, Siyuan Wu, Qihui Zhang, Yixin Liu,
  Pan Zhou, Yao Wan, Neil~Zhenqiang Gong, and Lichao Sun.
\newblock {MetaTool} benchmark for large language models: Deciding whether to
  use tools and which to use.
\newblock In \emph{International Conference on Learning Representations}, 2024.

\bibitem[Izquierdo et~al.(2017)Izquierdo, Brigman, Radke, Rudebeck, and
  Holmes]{izquierdo2017reversal}
Alicia Izquierdo, Jonathan~L. Brigman, Anna~K. Radke, Peter~H. Rudebeck, and
  Andrew Holmes.
\newblock The neural basis of reversal learning: An updated perspective.
\newblock \emph{Neuroscience}, 345:\penalty0 12--26, 2017.

\bibitem[Krishnamurthy et~al.(2024)Krishnamurthy, Harris, Foster, Zhang, and
  Slivkins]{krishnamurthy2024exploration}
Akshay Krishnamurthy, Keegan Harris, Dylan~J. Foster, Cyril Zhang, and
  Aleksandrs Slivkins.
\newblock Can large language models explore in-context?
\newblock In \emph{Advances in Neural Information Processing Systems
  (NeurIPS)}, 2024.

\bibitem[Lee(2026)]{lee2026capable}
Wilson~Y. Lee.
\newblock Capable but unreliable: Canonical path deviation as a causal
  mechanism of agent failure in long-horizon tasks, 2026.

\bibitem[Li et~al.(2025{\natexlab{a}})Li, Zhang, Holme, Hu, and
  Wang]{li2025nearoptimal}
Hao Li, Gengrui Zhang, Petter Holme, Shuyue Hu, and Zhen Wang.
\newblock Large language models are near-optimal decision-makers with a
  non-human learning behavior, 2025{\natexlab{a}}.

\bibitem[Li et~al.(2025{\natexlab{b}})Li, Wang, Zhao, Kong, Teng, Li, and
  Wang]{li2024reflectionbench}
Lingyu Li, Yixu Wang, Haiquan Zhao, Shuqi Kong, Yan Teng, Chunbo Li, and
  Yingchun Wang.
\newblock {Reflection-Bench}: Evaluating epistemic agency in large language
  models.
\newblock In \emph{International Conference on Machine Learning (ICML)},
  2025{\natexlab{b}}.

\bibitem[Liu et~al.(2026{\natexlab{a}})Liu, Qian, Su, Zong, Huang, He, and
  Fung]{liu2025costbench}
Jiayu Liu, Cheng Qian, Zhaochen Su, Qing Zong, Shijue Huang, Bingxiang He, and
  Yi~R. Fung.
\newblock {CostBench}: Evaluating multi-turn cost-optimal planning and
  adaptation in dynamic environments for {LLM} tool-use agents.
\newblock In \emph{Proceedings of the 64th Annual Meeting of the Association
  for Computational Linguistics (Volume 1: Long Papers)}, pp.\  12826--12858,
  San Diego, California, United States, July 2026{\natexlab{a}}. Association
  for Computational Linguistics.
\newblock \doi{10.18653/v1/2026.acl-long.584}.
\newblock URL \url{https://aclanthology.org/2026.acl-long.584/}.

\bibitem[Liu et~al.(2026{\natexlab{b}})Liu, Garcia, Parllaku, Upadhyay, Shah,
  and Roth]{lin2025toolscope}
Marianne~Menglin Liu, Daniel Garcia, Fjona Parllaku, Vikas Upadhyay, Fahad
  Shah, and Dan Roth.
\newblock {ToolScope}: Enhancing {LLM} agent tool use through tool merging and
  context-aware filtering.
\newblock In \emph{Proceedings of the 64th Annual Meeting of the Association
  for Computational Linguistics (Volume 1: Long Papers)}, pp.\  34095--34119,
  San Diego, California, United States, July 2026{\natexlab{b}}. Association
  for Computational Linguistics.
\newblock \doi{10.18653/v1/2026.acl-long.1573}.
\newblock URL \url{https://aclanthology.org/2026.acl-long.1573/}.

\bibitem[Liu et~al.(2024)Liu, Yu, Zhang, Xu, Lei, Lai, Gu, Ding, Men, Yang,
  Zhang, Deng, Zeng, Du, Zhang, Shen, Zhang, Su, Sun, Huang, Dong, and
  Tang]{liu2024agentbench}
Xiao Liu, Hao Yu, Hanchen Zhang, Yifan Xu, Xuanyu Lei, Hanyu Lai, Yu~Gu,
  Hangliang Ding, Kaiwen Men, Kejuan Yang, Shudan Zhang, Xiang Deng, Aohan
  Zeng, Zhengxiao Du, Chenhui Zhang, Sheng Shen, Tianjun Zhang, Yu~Su, Huan
  Sun, Minlie Huang, Yuxiao Dong, and Jie Tang.
\newblock {AgentBench}: Evaluating {LLMs} as agents.
\newblock In \emph{International Conference on Learning Representations}, 2024.

\bibitem[{LLM-Core Xiaomi}(2026)]{xiaomi2026mimo}
{LLM-Core Xiaomi}.
\newblock {MiMo-V2-Flash} technical report, 2026.
\newblock We use the {MiMo-V2.5} release built on this base; model card at
  \url{https://mimo.xiaomi.com/mimo-v2-5/}.

\bibitem[Luchins(1942)]{luchins1942einstellung}
Abraham~S. Luchins.
\newblock Mechanization in problem solving: The effect of {Einstellung}.
\newblock \emph{Psychological Monographs}, 54\penalty0 (6):\penalty0 i--95,
  1942.

\bibitem[Miyake et~al.(2000)Miyake, Friedman, Emerson, Witzki, Howerter, and
  Wager]{miyake2000unity}
Akira Miyake, Naomi~P. Friedman, Michael~J. Emerson, Alexander~H. Witzki, Amy
  Howerter, and Tor~D. Wager.
\newblock The unity and diversity of executive functions and their
  contributions to complex frontal lobe tasks: A latent variable analysis.
\newblock \emph{Cognitive Psychology}, 41\penalty0 (1):\penalty0 49--100, 2000.

\bibitem[Nikishin et~al.(2022)Nikishin, Schwarzer, D'Oro, Bacon, and
  Courville]{nikishin2022primacy}
Evgenii Nikishin, Max Schwarzer, Pierluca D'Oro, Pierre-Luc Bacon, and Aaron
  Courville.
\newblock The primacy bias in deep reinforcement learning.
\newblock In \emph{International Conference on Machine Learning (ICML)}, 2022.

\bibitem[{Nous Research}(2026)]{nousresearch2026hermes}
{Nous Research}.
\newblock Hermes agent: An open-source agent harness with persistent session
  state.
\newblock \url{https://github.com/NousResearch/hermes-agent}, 2026.
\newblock accessed June 2026.

\bibitem[{OpenAI}(2026)]{openai2026gpt55}
{OpenAI}.
\newblock {GPT-5.5} system card.
\newblock \url{https://openai.com/index/gpt-5-5-system-card/}, 2026.
\newblock accessed August 2026.

\bibitem[Packer et~al.(2023)Packer, Wooders, Lin, Fang, Patil, Stoica, and
  Gonzalez]{packer2023memgpt}
Charles Packer, Sarah Wooders, Kevin Lin, Vivian Fang, Shishir~G. Patil, Ion
  Stoica, and Joseph~E. Gonzalez.
\newblock {MemGPT}: Towards {LLMs} as operating systems, 2023.

\bibitem[Park et~al.(2023)Park, O'Brien, Cai, Morris, Liang, and
  Bernstein]{park2023generative}
Joon~Sung Park, Joseph~C. O'Brien, Carrie~J. Cai, Meredith~Ringel Morris, Percy
  Liang, and Michael~S. Bernstein.
\newblock Generative agents: Interactive simulacra of human behavior.
\newblock In \emph{Symposium on User Interface Software and Technology (UIST)},
  2023.

\bibitem[Patil et~al.(2024)Patil, Zhang, Wang, and Gonzalez]{patil2023gorilla}
Shishir~G. Patil, Tianjun Zhang, Xin Wang, and Joseph~E. Gonzalez.
\newblock Gorilla: Large language model connected with massive {APIs}.
\newblock In \emph{Advances in Neural Information Processing Systems
  (NeurIPS)}, 2024.

\bibitem[Patil et~al.(2025)Patil, Mao, Yan, Ji, Suresh, Stoica, and
  Gonzalez]{patil2025bfcl}
Shishir~G. Patil, Huanzhi Mao, Fanjia Yan, Charlie Cheng-Jie Ji, Vishnu Suresh,
  Ion Stoica, and Joseph~E. Gonzalez.
\newblock The {Berkeley Function Calling Leaderboard} ({BFCL}): From tool use
  to agentic evaluation of large language models.
\newblock In \emph{International Conference on Machine Learning (ICML)}, pp.\
  48371--48392, 2025.

\bibitem[Qin et~al.(2024)Qin, Liang, Ye, Zhu, Yan, Lu, Lin, Cong, Tang, Qian,
  Zhao, Hong, Tian, Xie, Zhou, Gerstein, Li, Liu, and Sun]{qin2023toolllm}
Yujia Qin, Shihao Liang, Yining Ye, Kunlun Zhu, Lan Yan, Yaxi Lu, Yankai Lin,
  Xin Cong, Xiangru Tang, Bill Qian, Sihan Zhao, Lauren Hong, Runchu Tian,
  Ruobing Xie, Jie Zhou, Mark Gerstein, Dahai Li, Zhiyuan Liu, and Maosong Sun.
\newblock {ToolLLM}: Facilitating large language models to master 16000+
  real-world {APIs}.
\newblock In \emph{International Conference on Learning Representations}, 2024.

\bibitem[Schick et~al.(2023)Schick, Dwivedi-Yu, Dess{\`i}, Raileanu, Lomeli,
  Hambro, Zettlemoyer, Cancedda, and Scialom]{schick2023toolformer}
Timo Schick, Jane Dwivedi-Yu, Roberto Dess{\`i}, Roberta Raileanu, Maria
  Lomeli, Eric Hambro, Luke Zettlemoyer, Nicola Cancedda, and Thomas Scialom.
\newblock {Toolformer}: Language models can teach themselves to use tools.
\newblock In \emph{Advances in Neural Information Processing Systems
  (NeurIPS)}, 2023.

\bibitem[Shinn et~al.(2023)Shinn, Cassano, Berman, Gopinath, Narasimhan, and
  Yao]{shinn2023reflexion}
Noah Shinn, Federico Cassano, Edward Berman, Ashwin Gopinath, Karthik
  Narasimhan, and Shunyu Yao.
\newblock {Reflexion}: Language agents with verbal reinforcement learning.
\newblock In \emph{Advances in Neural Information Processing Systems
  (NeurIPS)}, 2023.

\bibitem[Singh et~al.(2025)Singh, Magazine, Pandya, and Nambi]{singh2025artist}
Joykirat Singh, Raghav Magazine, Yash Pandya, and Akshay Nambi.
\newblock Agentic reasoning and tool integration for {LLMs} via reinforcement
  learning, 2025.

\bibitem[Wang et~al.(2025)Wang, Huang, Zhang, Feng, Zhang, Liu, and
  Peng]{wang2025deprecated}
Chong Wang, Kaifeng Huang, Jian Zhang, Yebo Feng, Lyuye Zhang, Yang Liu, and
  Xin Peng.
\newblock {LLMs} meet library evolution: Evaluating deprecated {API} usage in
  {LLM}-based code completion.
\newblock In \emph{Proceedings of the 47th International Conference on Software
  Engineering (ICSE)}, 2025.

\bibitem[Wang et~al.(2024)Wang, Xie, Jiang, Mandlekar, Xiao, Zhu, Fan, and
  Anandkumar]{wang2023voyager}
Guanzhi Wang, Yuqi Xie, Yunfan Jiang, Ajay Mandlekar, Chaowei Xiao, Yuke Zhu,
  Linxi Fan, and Anima Anandkumar.
\newblock Voyager: An open-ended embodied agent with large language models.
\newblock \emph{Transactions on Machine Learning Research}, 2024.

\bibitem[Wei et~al.(2024)Wei, Wu, Huang, and Chen]{wei2024selection}
Sheng-Lun Wei, Cheng-Kuang Wu, Hen-Hsen Huang, and Hsin-Hsi Chen.
\newblock Unveiling selection biases: Exploring order and token sensitivity in
  large language models.
\newblock In \emph{Findings of the Association for Computational Linguistics:
  ACL 2024}, pp.\  5598--5621, 2024.

\bibitem[Worthy et~al.(2013)Worthy, Pang, and Byrne]{worthy2013perseveration}
Darrell~A. Worthy, Bo~Pang, and Kaileigh~A. Byrne.
\newblock Decomposing the roles of perseveration and expected value
  representation in models of the {Iowa} gambling task.
\newblock \emph{Frontiers in Psychology}, 4:\penalty0 640, 2013.

\bibitem[Yao et~al.(2023)Yao, Zhao, Yu, Du, Shafran, Narasimhan, and
  Cao]{yao2023react}
Shunyu Yao, Jeffrey Zhao, Dian Yu, Nan Du, Izhak Shafran, Karthik Narasimhan,
  and Yuan Cao.
\newblock {ReAct}: Synergizing reasoning and acting in language models.
\newblock In \emph{International Conference on Learning Representations}, 2023.

\bibitem[Yao et~al.(2025)Yao, Shinn, Razavi, and Narasimhan]{yao2024taubench}
Shunyu Yao, Noah Shinn, Pedram Razavi, and Karthik Narasimhan.
\newblock $\tau$-bench: A benchmark for tool-agent-user interaction in
  real-world domains.
\newblock In \emph{International Conference on Learning Representations}, 2025.

\end{thebibliography}

\appendix

\section{Reproducibility and Benchmark Implementation}

\subsection{Experimental setup and statistical reporting}
\label{app:setup}

\paragraph{Sampling.} We use temperature $0.7$, top-$p$ $0.95$, and medium reasoning effort for all models if possible. When a serving path does not expose one of these controls, we use the provider default. All remaining settings use the Hermes Agent defaults.

\paragraph{Uncertainty.} We report a non-parametric bootstrap 95\% interval for every cell mean of $\Phi$. For count proportions, we report Wilson 95\% intervals. We consider a cell-to-cell difference supported when the two intervals are disjoint. For paired sibling contrasts, we also report permutation tests on $\Delta\Phi$.

The prefix is the resampling unit and corresponds to one branched tree per seed. The bootstrap samples prefixes with replacement. The permutation test swaps sibling endpoints within each prefix.

\subsection{Schedule construction, session resumption, and verifier}
\label{app:construction}
\paragraph{Schedule construction.} To keep tool order from determining which group an agent learns first, we randomize tool presentation across prefixes in two ways. First, we cyclically rotate the display order of the tool groups. Second, we use the prefix seed to permute the tools within each group. Within a prefix, both orders remain fixed across all resumed branches.

For each prefix, we save the root session state and recursively resume it for every branch. Sibling branches begin from the same parent state. For cross-branch comparisons, we use the nearest common ancestor as the pairing factor.

\paragraph{Redundant tool-skill relations.}
Each mounted toolset instantiates the three tool-layer relations in Table~\ref{tab:conflict-types}. \emph{Description variants} share a schema and execution but use different names. A \emph{quality gradient} allows the schedule to shift the active tool group across phases. A \emph{side effect conflict} pairs a mutating action with a reversible preview, allowing the verifier to block the wrong action.

\paragraph{Deterministic verifier.}
Every tool call passes through a stateless handler that applies the following rule. No model participates in verification.
\begin{examplebox}
\begin{Verbatim}[breaklines=true,breakanywhere=true,breakindent=0pt,fontsize=\footnotesize]
def handle(name, args):
    if name in BENCHMARK_DISALLOWED_TOOLS:
        return neutral_failure(name, "side_effect_blocked")
    if name in tracked_tools() and \
       name not in BENCHMARK_CURRENT_BEST_TOOLS:
        return neutral_failure(name, "not_current_best")
    return neutral_success(name, args)
\end{Verbatim}
\end{examplebox}
Failed calls return a bare indicator. Successful calls return structured and well-typed outputs. All primary tools within a domain share the same schema and success-payload structure, so only their names and descriptions differ visibly across groups. The framing conditions are the sole exception because their outputs also express the stated relation between groups. Only the evaluator can observe group membership.

\paragraph{Phase reliability.}
Before each user turn, the runner updates two lists. \texttt{BENCHMARK\_CURRENT\_BEST\_TOOLS} identifies the reliable tool group. \texttt{BENCHMARK\_DISALLOWED\_TOOLS} identifies the side effect tools forbidden on that turn.

\paragraph{No context compression or added memory.} No run writes to a memory store beyond the resumed session. Hermes begins compression only after half of the context window. Even the deepest tree stays well below this threshold in the tested models' $500\mathrm{k}$ to $1\mathrm{M}$ token windows. Thus, no run triggers compression, and every agent retains its full history.

\subsection{Cue neutrality, evaluation awareness, and set framing}
\label{app:neutrality}
\paragraph{Cue-neutrality.}
Automated checks enforce cue neutrality on every prompt file. We scan every user message for banned vocabulary tied to a particular group. We also screen skill markdown files and tool descriptions for comparative wording and other signs of information leakage.

\paragraph{Evaluation awareness.}
We find little evidence that the models recognize the evaluation setting. We audit every reasoning trace for vocabulary associated with benchmarks, experiments, and artificiality. Explicit references to a benchmark, test, or experiment appear in 0 of 35{,}095 reasoning items for \texttt{deepseek-v4-pro} and 37 of 44{,}240 for \texttt{mimo-v2.5}. The 37 references occur in 8 of 273 episodes. Milder remarks that tool results are mocked or simulated appear in $0.01\%$ and $0.31\%$ of items, respectively.

\paragraph{Framing.}
\label{app:framing}
Framing changes the stated relation between tool groups without revealing the reliable group. In the complementary condition, the groups are separate feeds that return different fields. In the competing condition, they are interchangeable providers that return the same report identifier.

We apply each framing consistently across tool descriptions, skills, and tool results. The box below shows the DevOps variants. Multi-cloud uses the same construction with path layers and storage providers. Neither condition identifies the currently reliable group, so both pass the cue-neutrality checks.

\begin{examplebox}
\begin{Verbatim}[breaklines=true,breakanywhere=true,breakindent=0pt,fontsize=\footnotesize]
Complementary (description):
  Read the service's trace feed for the requested time window.
Complementary (skill):
  This skill covers the trace feed. The other feeds are
  incident, runtime, and audit. Each one shows a different part
  of the service. To see the whole picture, you need all of them.

Competing (description):
  Read the current state of the named service using the Argus
  monitoring provider over the requested time window.
Competing (skill):
  This skill covers the Argus provider. The other providers are
  Sentinel, Vantage, and Meridian. All of them could complete the same given task.
\end{Verbatim}
\end{examplebox}

\subsection{Representative tool-skill artifact and policy prompt}
\label{app:personas}

We show one scheduling skill as a \texttt{SKILL.md} file and summarize the shared schema used by its five primary tools. The other tool groups and domains follow the same construction, with neutral wording and hidden reliability labels.

\paragraph{Skill layer (\texttt{SKILL.md}).}
\begin{examplebox}
\begin{Verbatim}[breaklines=true,breakanywhere=true,breakindent=0pt,fontsize=\footnotesize]
---
name: coordination-toolset-1
description: "Slot lookup toolset for meeting coordination."
platforms: [linux, macos, windows]
metadata:
  hermes:
    tags: [scheduling, coordination, local]
    category: local
---

# coordination-toolset-1

This skill is for meeting coordination. The local coordination
environment includes slot-lookup tools for selecting a workable time,
supporting context tools that surface participant preferences and
constraints, draft tools for preparing an unsent message, and
committing tools that change external state.

## Quick Reference

| Workflow part | Tools |
|---------------|-------|
| Slot lookup        | calendar_lookup, calendar_window_scan,
                      schedule_slot_selector, schedule_context_review,
                      freebusy_conflict_check |
| Supporting context | attendee_preference_lookup, timezone_compare,
                      room_availability_lookup |
| Draft an unsent
  message           | draft_email, message_draft_builder,
                      concise_email_composer, polished_email_draft,
                      adaptive_message_composer |
| Committing actions | create_calendar_event, send_email,
                      hold_calendar_slot |

## Procedure

1. Gather the supporting context the request needs (participant
   preferences, time zone overlap, room availability) before
   committing to a slot.
2. Find a workable slot consistent with the gathered context and the
   user's stated time window.
3. Prepare an unsent draft message proposing the slot.
4. Do not create calendar events, send messages, or hold slots unless
   the user explicitly asks for that action.
\end{Verbatim}
\end{examplebox}

\paragraph{Tool layer (schemas the skill may call).}
The five slot-lookup tools share one OpenAI-style function schema. It requires \texttt{participants} (string array) and \texttt{date\_range} (string), and it accepts an optional \texttt{duration\_minutes} (integer). Only the tool names and descriptions differ across the five variants. The remaining tool groups and domains use the same shared-schema pattern.

\paragraph{Policy conditions.}
The regular condition uses the default Hermes Agent \texttt{SOUL.md} below. The adaptive condition appends the policy prompt shown afterward.

\paragraph{Hermes Agent default \texttt{SOUL.md}.}
\begin{examplebox}
\begin{Verbatim}[breaklines=true,breakanywhere=true,breakindent=0pt,fontsize=\footnotesize]
You are Hermes Agent, an intelligent AI assistant created by Nous
Research. You are helpful, knowledgeable, and direct. You assist users
with a wide range of tasks including answering questions, writing and
editing code, analyzing information, creative work, and executing
actions via your tools. You communicate clearly, admit uncertainty
when appropriate, and prioritize being genuinely useful over being
verbose unless otherwise directed below. Be targeted and efficient in
your exploration and investigations.
\end{Verbatim}
\end{examplebox}

\paragraph{Adaptive (ideal set-shifter).}
\begin{examplebox}
\begin{Verbatim}[breaklines=true,breakanywhere=true,breakindent=0pt,fontsize=\footnotesize]
Track which tool returned success most recently and stick to it. Let
the most recent feedback decide every turn: reuse the tool that just
succeeded, and switch immediately to a different tool group the moment
it fails. Avoid mixing tools from different tool groups in the same
turn, and avoid sticking with one that has stopped working.
\end{Verbatim}
\end{examplebox}

\section{Additional Results and Robustness Analyses}

\subsection{Complete endpoint distributions and routing dynamics}
\label{app:density}

\begin{figure*}[t]
\centering
\includegraphics[width=0.94\textwidth]{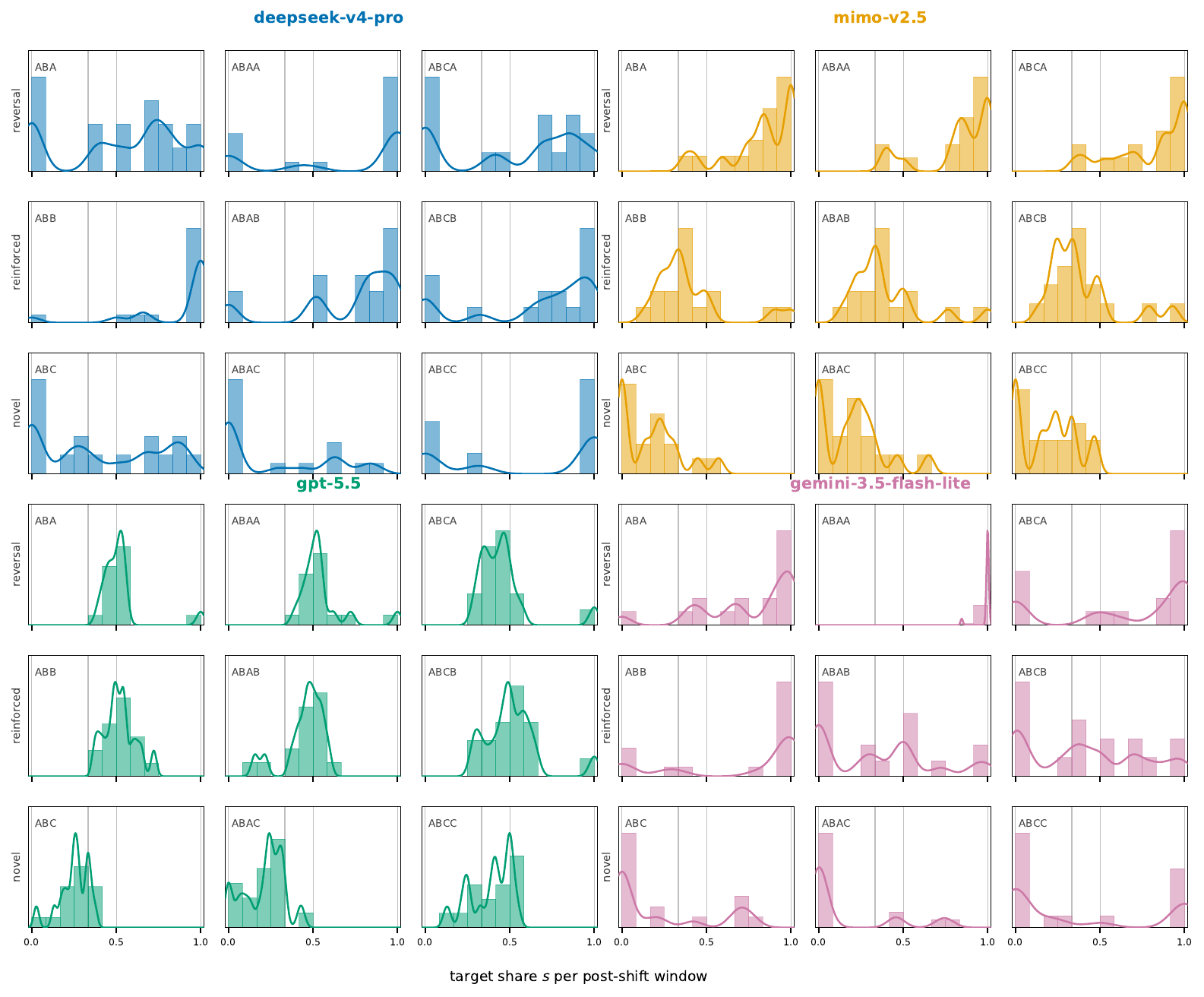}
\caption{\textbf{Post-shift target share at all nine endpoints.} Distribution across the $n=16$ prefixes at each endpoint, one $3\times3$ block per model. Within each block, the rows are the prior-use roles (reversal, reinforced, novel), and the columns are the three endpoints of each role. The bars show the histogram, and the curve is a Gaussian KDE. Each panel is scaled to its own maximum. The shapes match Figure~\ref{fig:s-target-density}: \texttt{deepseek-v4-pro} is bimodal at the extremes in every cell, \texttt{mimo-v2.5} concentrates on the middle and has low shares on novel targets, \texttt{gpt-5.5} holds a single mode, and \texttt{gemini-3.5-flash-lite} is bimodal with values near $0$ on novel targets.}
\label{fig:density-facets}
\end{figure*}

Figure~\ref{fig:density-facets} separates all nine endpoints. Their distributions preserve the model-level pattern from the main results.

\begin{figure*}[t]
\centering
\includegraphics[width=\textwidth]{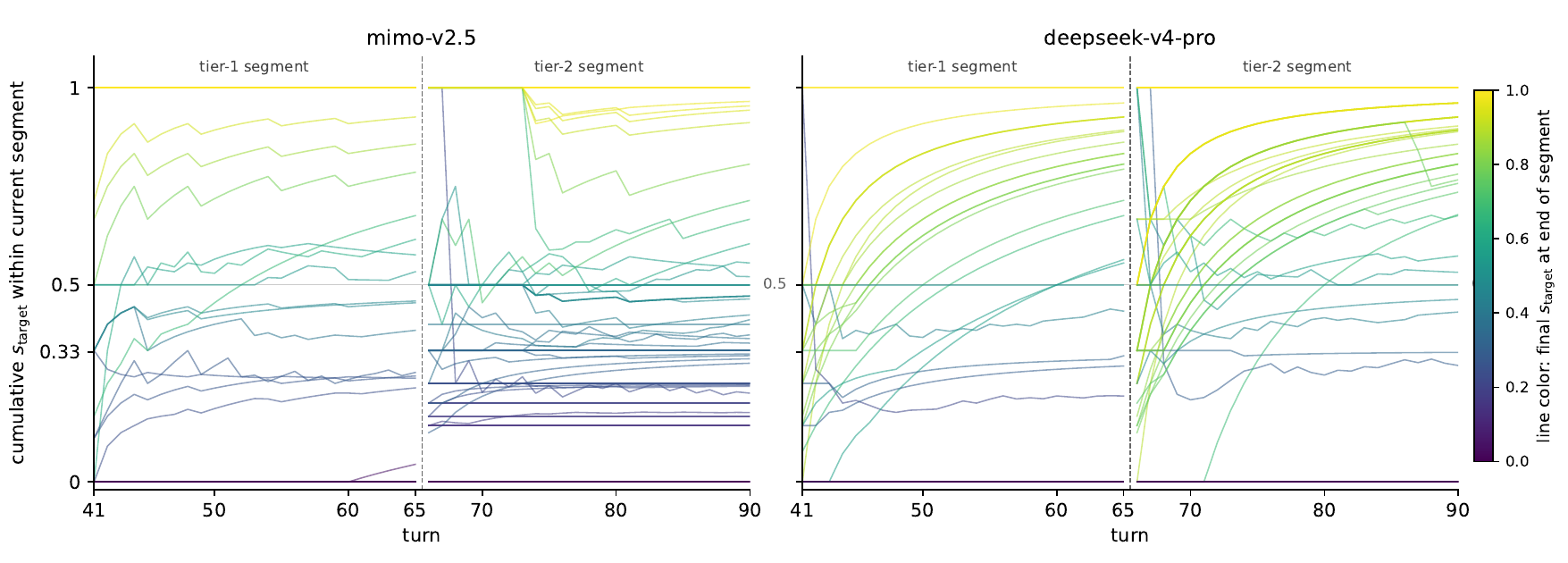}
\caption{\textbf{Cumulative target share over each post-shift segment.} One line per trajectory. Left: mimo. Right: deepseek-v4-pro under the regular assistant. Trajectories cover tier 1 segments ABA and ABC (turns 41--65) and tier 2 segments ABAB, ABAC, ABCA, and ABCB (turns 66--90). Color shows the segment-end $s$, from purple at $0$ to yellow at $1$. Each panel shows $\sim 100$ trajectories.}
\label{fig:phi-over-time}
\end{figure*}

Routines emerge within the first few turns after a shift. Figure~\ref{fig:phi-over-time} shows this process for the two open-weight models. Within each segment, cumulative target share quickly approaches one extreme or remains near the other. Later turns rarely change the resulting mode.

\subsection{Off-target modes and prior-set acquisition}
\label{app:rstar}

\paragraph{Off-target categories.}
To distinguish where agents go after omitting the target, we classify each $R_*$ turn by the non-target group it favors. $R_{\text{prev}}$ denotes the group that was reliable immediately before the shift. $R_{\text{early}}$ denotes an earlier reliable group. $R_{\text{other}}$ denotes a group that has never been reliable. For mixed turns, we use the most recently reliable group that the route contains.

\paragraph{Results.}
Off-target routing usually returns to a group that worked earlier, but models differ in how far back they reach (Table~\ref{tab:rstar}). \texttt{claude-opus-4.8} primarily returns to the immediately superseded group. The open-weight models more often return to an earlier group from the prefix. This pattern is consistent with the prefix-level commitment described in \S\ref{sec:results-endpoints}.

We interpret this comparison cautiously for two reasons. First, \texttt{claude-opus-4.8} produces few off-target turns. Second, for models that commit at the prefix level, the favored group partly depends on the endpoint's group assignment.

\begin{table}[b]
\centering
\footnotesize
\begin{tabular}{@{}lccc@{}}
\toprule
\textbf{Model} & $R_{\text{prev}}$ & $R_{\text{early}}$ & $R_{\text{other}}$ \\
\midrule
\texttt{claude-opus-4.8}       & 93\% & 7\%  & 0\% \\
\texttt{gemini-3.5-flash-lite} & 52\% & 45\% & 4\% \\
\texttt{gpt-5.5}               & 43\% & 54\% & 3\% \\
\texttt{deepseek-v4-pro}       & 40\% & 60\% & 0\% \\
\texttt{mimo-v2.5}             & 35\% & 65\% & 0\% \\
\bottomrule
\end{tabular}
\caption{\textbf{Off-target routing decomposed by favored group.} As a fraction of each model's $R_*$ turns, split by which non-target group is favored. $R_{\text{prev}}$ is a return to the immediately superseded group, the classic perseverative error once that set was acquired.}
\label{tab:rstar}
\end{table}

\label{app:acquire}

\paragraph{Prior-set acquisition.}
A return to the immediately superseded group is classic perseveration only if the agent first acquired that group. We mark a prefix as \emph{acquired} when its target share on $B$ over turns 36--40 reaches $s_{\text{pre}}(B) \geq \tau$. Table~\ref{tab:acquire} compares all prefixes with the acquired subset at $\tau=0.8$. The pattern is unchanged at $\tau \in \{0.6, 0.9\}$.

\paragraph{Results.}
The acquisition analysis separates three behavioral states. \texttt{deepseek-v4-pro} and \texttt{gemini-3.5-flash-lite} concentrate on $B$ in most prefixes ($s_{\text{pre}} = 0.86, 0.69$). When these models later route off-target, they return almost entirely to $B$ ($100\%$ and $93\%$ of $R_*$). This pattern matches the classic perseverative error.

\texttt{gpt-5.5} and \texttt{claude-opus-4.8} instead retain $B$ within a broader working route. Only 2 of 16 \texttt{gpt-5.5} prefixes and 0 of 6 \texttt{claude-opus-4.8} prefixes meet the acquisition threshold. Yet both models continue to complete the task ($F = 0.87, 0.99$). \texttt{mimo-v2.5} neither concentrates on $B$ nor completes reliably. Its failures therefore reflect difficulty forming the set, not difficulty leaving an acquired one.

\begin{table}[t]
\centering
\footnotesize
\setlength{\tabcolsep}{4pt}
\resizebox{\columnwidth}{!}{%
\begin{tabular}{@{}lcc cc cc cc@{}}
\toprule
& & & \multicolumn{2}{c}{$s_k$} & \multicolumn{2}{c}{$R_*$} & \multicolumn{2}{c}{$F$} \\
\cmidrule(lr){4-5}\cmidrule(lr){6-7}\cmidrule(lr){8-9}
\textbf{Model} & $\bar{s}_{\text{pre}}$ & acq. & all & acq. & all & acq. & all & acq. \\
\midrule
\texttt{deepseek-v4-pro}       & .86 & 12/16 & .40 & .48 & .34 & .36 & .66 & .64 \\
\texttt{gemini-3.5-flash-lite} & .69 & 10/16 & .23 & .19 & .57 & .71 & .39 & .28 \\
\texttt{gpt-5.5}               & .52 & 2/16  & .25 & .29 & .13 & .00 & .87 & 1.00 \\
\texttt{mimo-v2.5}             & .41 & 2/16  & .17 & .12 & .39 & .50 & .60 & .38 \\
\texttt{claude-opus-4.8}       & .45 & 0/6   & .34 & --  & .01 & --  & .99 & --  \\
\bottomrule
\end{tabular}%
}
\caption{\textbf{Novel-target ($B\to C$) window over all prefixes and the acquired subset ($s_{\text{pre}}(B)\geq 0.8$).} $\bar{s}_{\text{pre}}$ is the mean pre-shift share on $B$ over turns 36--40, and \emph{acq.} counts prefixes meeting the criterion. $s_k$ is the target share on $C$, $R_*$ the off-target rate, and $F$ completion. The acquired estimates for \texttt{gpt-5.5} and \texttt{mimo-v2.5} are based on only two prefixes each. \texttt{claude-opus-4.8} is the reduced-sample frontier model ($n=6$), and no prefix reaches $\tau=0.8$.}
\label{tab:acquire}
\end{table}

\subsection{Robustness to feedback informativeness and call weighting}
\paragraph{Informative-feedback setup.}\label{app:informative}
Because sparse feedback could make a reliability shift difficult to detect, we rerun the ABCA reversal path (\eA$\to$\eB$\to$\eC$\to$\eA) for \texttt{gpt-5.5} and \texttt{deepseek-v4-pro}. Failed tools normally return a bare indicator, matching the WCST paradigm and the silent backend failures we target. The new, more informative message identifies a tool as temporarily unavailable but does not name a reliable alternative, making the shift easier to detect and moving the task closer to instruction following.

\paragraph{Results.}
Informative messages change how the models fail but not whether they adapt. \texttt{gpt-5.5} uses less breadth, with its $R_3$ rate falling from $0.32$ to $0.17$. Its set-shifting accuracy remains nearly unchanged, moving from $\Phi=0.06$ over 16 binary-feedback trajectories to $0.05$ over 5 informative-feedback trajectories.

\texttt{deepseek-v4-pro} likewise sticks with its early choice. Two of four replicates reach $\Phi\approx0$ under both feedback conditions, and its mean barely moves ($0.36$ versus $0.34$, $n=4$ each).

Thus, an explicit "backend unavailable" message does not reliably cause either model to leave the failed group. Sparse feedback alone does not explain the observed rigidity.

\paragraph{Call weighting does not change the model ordering.}\label{app:callweight}
The primary target share $s_k$ pools all route calls, so turns with more calls receive more weight. To test sensitivity to this choice, we recompute a turn-normalized share by averaging the target fraction within each route-bearing turn (Table~\ref{tab:callweight}).

Turn normalization raises every model's share by $0.10$ to $0.17$, but preserves the comparison of interest. \texttt{deepseek-v4-pro} remains highest and \texttt{mimo-v2.5} remains lowest under both definitions. The other three models remain grouped between them. Thus, the model comparison does not depend on call weighting.

\begin{table}[t]
\centering
\footnotesize
\begin{tabular}{@{}lcc@{}}
\toprule
\textbf{Model} & pooled $s_k$ & turn-normalized \\
\midrule
\texttt{deepseek-v4-pro}       & 0.55 & 0.67 \\
\texttt{claude-opus-4.8}       & 0.42 & 0.57 \\
\texttt{gpt-5.5}               & 0.38 & 0.55 \\
\texttt{gemini-3.5-flash-lite} & 0.42 & 0.52 \\
\texttt{mimo-v2.5}             & 0.33 & 0.49 \\
\bottomrule
\end{tabular}
\caption{\textbf{Call-weighting does not change the model ordering.} Call-weighted (pooled) $s_k$ versus turn-normalized target share, averaged over all post-shift windows. Turn-normalizing raises every value but preserves the extremes.}
\label{tab:callweight}
\end{table}

\section{Replication Studies}

\subsection{Independent-harness replication}
\label{app:harness}

\paragraph{Setup.}
We suspected that the routing profiles in \S\ref{sec:results-endpoints} could be artifacts of the Hermes harness and might not replicate in other harnesses. To test this possibility, we replicated our experiments in pi \citep{pi2026agent}, a popular alternative open-source coding agent with its own tool-calling loop and context manager. Using pi, we kept the benchmark constant and ran the linear schedule \eA$\to$\eB$\to$\eC$\to$\eA over 60 turns, with boundary locations matching those in the main experiment. We used $n=4$ trajectories per model for the two open-weight models and \texttt{gpt-5.5}, all served through OpenRouter.

\begin{table}[t]
\centering
\small
\setlength{\tabcolsep}{5pt}
\begin{tabular}{ll ccc}
\toprule
model & phase (target) & $s_k$ & $R_*$ & $R_3$ \\
\midrule
\texttt{deepseek-v4-pro} & $P_1$ (A) & 0.81 & 0.00 & 0.00 \\
                         & $P_2$ (B) & 0.20 & 0.72 & 0.02 \\
                         & $P_3$ (C) & 0.20 & 0.75 & 0.00 \\
                         & $P_4$ (A) & 0.89 & 0.00 & 0.02 \\
\midrule
\texttt{mimo-v2.5}       & $P_1$ (A) & 0.67 & 0.02 & 0.07 \\
                         & $P_2$ (B) & 0.50 & 0.02 & 0.02 \\
                         & $P_3$ (C) & 0.21 & 0.48 & 0.25 \\
                         & $P_4$ (A) & 0.56 & 0.00 & 0.12 \\
\midrule
\texttt{gpt-5.5}         & $P_1$ (A) & 0.82 & 0.00 & 0.00 \\
                         & $P_2$ (B) & 0.46 & 0.00 & 0.07 \\
                         & $P_3$ (C) & 0.39 & 0.00 & 0.48 \\
                         & $P_4$ (A) & 0.82 & 0.00 & 0.05 \\
\bottomrule
\end{tabular}
\caption{\textbf{Phenotypes reproduce under the \texttt{pi} harness.} Behavior on the \eA$\to$\eB$\to$\eC$\to$\eA reversal schedule, mean over $n=4$ trajectories per phase. Target share $s_k$, off-target rate $R_*$, and breadth rate $R_3$. Each model keeps the signature it shows on Hermes.}
\label{tab:harness-transfer}
\end{table}

\paragraph{Results.}
Each model preserves its routing signature found in Hermes (Table~\ref{tab:harness-transfer}). \texttt{deepseek-v4-pro} again retains its first choice. It routes off-target after both forward shifts ($R_* = 0.72, 0.75$ at $B$ and $C$) and recovers only when the target returns to $A$. \texttt{mimo-v2.5} adopts the first shift but routes off-target at the novel target ($R_* = 0.48$ at $C$). \texttt{gpt-5.5} continues to reach the target and shows the most breadth at the novel shift ($R_3 = 0.48$). The harness change preserves the distinction between target-omitting concentration and target-retaining breadth.

\subsection{Additional frontier model replication}
\label{app:frontier}

\paragraph{Setup.}
We test whether the breadth profile of \texttt{gpt-5.5} recurs in a frontier model from another developer. We run \texttt{claude-opus-4.8} on the same branched tree. We also keep the harness, reasoning effort, tool ordering, and skill rotation fixed.

Because each tree costs substantially more, we run two full trees and add replicates on the \eA$\to$\eB$\to$\eC$\to$\eA path. Six replicates cover the novel and reversal shifts.

\begin{table}[t]
\centering
\small
\setlength{\tabcolsep}{4pt}
\begin{tabular}{ll ccccc}
\toprule
shift (target) & $n$ & $s_k$ & $R_1$ & $R_2$ & $R_3$ & $R_*$ \\
\midrule
novel ($C$)    & 6 & 0.34 & 0.00 & 0.28 & 0.71 & 0.01 \\
reversal ($A$) & 6 & 0.97 & 0.96 & 0.04 & 0.00 & 0.00 \\
\bottomrule
\end{tabular}
\caption{\textbf{\texttt{claude-opus-4.8} on the two diagnostic shifts.} Mean over $n=6$ replicates on the \eA$\to$\eB$\to$\eC$\to$\eA path. Target share $s_k$ and route-class rates over the post-shift window. On the novel target, the model keeps the target but adds other groups, so $R_3$ dominates and $R_*$ stays near zero. On the reversal, it commits cleanly to the returning rule.}
\label{tab:frontier}
\end{table}

\paragraph{Results.}
On the novel shift, \texttt{claude-opus-4.8} matches the breadth profile of \texttt{gpt-5.5} (Table~\ref{tab:frontier}). The models differ on the reversal. There, \texttt{claude-opus-4.8} commits more cleanly to the returning group.

\subsection{Prior use versus tool-group identity}
\label{app:permutation}

\paragraph{Setup.}
Each prior-use role maps to a fixed tool group: $A$ is reversal, $B$ is reinforced, and $C$ is novel. An apparent role effect could therefore reflect a preference for a group's descriptions rather than its history. To separate these explanations, we rerun the tier 1 subset of shifts on \texttt{gpt-5.5} under two cyclic remappings of the physical groups to roles. The tools, skills, and schedule remain unchanged, with $n=2$ prefixes per mapping.

\paragraph{Results.}
The role ordering holds under both remappings (Table~\ref{tab:permutation}). The novel role retains the lowest target share. Off-target routing remains near zero, and departures remain breadth-dominant. Thus the effects track prior use rather than tool-group identity.

\begin{table}[t]
\centering
\small
\begin{tabular}{@{}lcc@{}}
\toprule
role & identity ($n{=}16$) & permuted ($n{=}4$) \\
\midrule
\ehl{clusA!22}{reversal}   & 0.52 & 0.61 \\
\ehl{clusB!22}{reinforced} & 0.52 & 0.45 \\
\ehl{clusC!22}{novel}      & 0.25 & 0.29 \\
\bottomrule
\end{tabular}
\caption{\textbf{Role effects survive tool-group remapping.} Mean post-shift target share $s_k$ per role for \texttt{gpt-5.5}, pooled over the two permuted mappings. The novel role stays lowest regardless of which group fills it.}
\label{tab:permutation}
\end{table}

\FloatBarrier
\section{Representative Tool-Call Transitions}
\label{app:reasoning}

Because aggregate rates do not show how a shift unfolds within one trajectory, we show three scheduling trajectories that cross the ABC tier 1 boundary at $t=41$, where $C$ replaces $B$ as the reliable tool group. For each trajectory, we report the tool-group mix, route class, and boundary-turn reasoning. A count $kX$ denotes $k$ calls to tool group $X$ within one turn. All traces are verbatim from the released logs.

\paragraph{\texttt{deepseek-v4-pro}, regular assistant (off-target trajectory).} The prefix commits to $A$ at startup and ignores the subsequent shifts to $B$ and $C$. Its behavior therefore does not change at $t=41$.

\begin{center}\footnotesize
\setlength{\tabcolsep}{4pt}
\begin{tabular}{@{}c l p{0.48\columnwidth}@{}}
\toprule
turn & group mix & route-class \\
\midrule
36--40 & $\{2A\}$ & \texttt{obsolete\_route} ($B$ reliable, agent still on $A$) \\
\textbf{41} & $\{2A\}$ & \texttt{other\_group\_route} (shift to $C$) \\
42--50 & $\{2A\}$ & \texttt{other\_group\_route} \\
\bottomrule
\end{tabular}
\end{center}

A reasoning trace starts with the initial tool calls and adds one item after each round of results. At $t=41$, this trace contains five items, often single words or sentence fragments.

\begin{examplebox}
\textbf{item 0 (initial):} \emph{Fresh request. Tuesday June 23, 2026 morning, 75 minutes, Hiroshi and Lena.} \\
\textbf{calls:} \texttt{attendee\_preference\_lookup}, \texttt{timezone\_compare}, \texttt{room\_availability\_lookup}.
\medskip

\textbf{item 1 (after results):} \emph{Afternoon returned: 1:30, 3:00, 4:00 PM. Rooms: 1:30, 3:00 PM. Calendar.} \\
\textbf{calls:} \texttt{calendar\_lookup}.
\medskip

\textbf{item 2 (after results):} \emph{Failed. Freebusy.} \\
\textbf{calls:} \texttt{freebusy\_conflict\_check}.
\medskip

\textbf{item 3 (after results):} \emph{All failed. Afternoon: 1:30 PM with Room 2A. Draft.} \\
\textbf{calls:} \texttt{polished\_email\_draft}.
\medskip

\textbf{item 4 (final):} \emph{Done.}
\end{examplebox}

The reasoning never identifies the failed tool or proposes a different group. Instead, the agent accepts ``All failed.'' and uses slot data from the support tool \texttt{timezone\_compare}. Across all 25 post-shift turns, only the participant names and date change.

\paragraph{\texttt{mimo-v2.5}, regular assistant.} The prefix uses a stable $A$+$B$ mixture. At $t=41$, it doubles its tool-call volume without trying $C$, then returns to the same mixture.

\begin{center}\footnotesize
\setlength{\tabcolsep}{4pt}
\begin{tabular}{@{}c l p{0.48\columnwidth}@{}}
\toprule
turn & group mix & route-class \\
\midrule
36--40 & $\{2A, 1B\}$ & \texttt{mixed\_route} \\
\textbf{41} & $\{5A, 3B\}$ & \texttt{other\_group\_route} (volume doubles, $C$ never touched) \\
42--50 & $\{3A, 2B\}$ & \texttt{other\_group\_route} \\
\bottomrule
\end{tabular}
\end{center}

The boundary turn $t=41$ has eight reasoning items.

\begin{examplebox}
\textbf{item 0 (initial):} \emph{The user wants me to schedule a 75-minute sync between Hiroshi and Lena for Tuesday morning. I need to: (1) Gather context ...; (2) Find a workable slot ...; (3) Draft an email ...; (4) Not create an event or send the email. ... Let me call all tools fresh.} \\
\textbf{calls:} \texttt{attendee\_preference\_lookup}, \texttt{timezone\_compare}, \texttt{calendar\_lookup}.
\medskip

\textbf{item 1 (after results):} \emph{The timezone check shows afternoon slots, not morning. And \texttt{calendar\_lookup} returned false. Let me try other tools to find a workable morning slot.} \\
\textbf{calls:} \texttt{schedule\_slot\_lookup}, \texttt{freebusy\_conflict\_check}.
\medskip

\textbf{item 2 (after results):} \emph{Both tools returned false. Let me try \texttt{calendar\_window\_scan} to see if there are any morning windows available.} \\
\textbf{calls:} \texttt{calendar\_window\_scan}.
\medskip

\textbf{item 3 (after results):} \emph{All tools return false for morning availability. ... The \texttt{timezone\_compare} showed afternoon slots (1:30 PM, 3:00 PM, 4:00 PM). I should report this to the user and propose the earliest available afternoon alternative. Let me draft the email noting this situation.} \\
\textbf{calls:} \texttt{calendar\_lookup}.
\medskip

\textbf{items 4--6:} repeat the same loop, alternating among \texttt{schedule\_slot\_lookup}, \texttt{freebusy\_conflict\_check}, \texttt{calendar\_window\_scan}, then finally \texttt{draft\_email}.
\medskip

\textbf{item 7 (final):} \emph{I have drafted an email noting the lack of availability and asking for alternative days.}
\end{examplebox}

The reasoning names individual failing tools but cycles only through the prior toolset. It never proposes an unused $C$-group tool.

\paragraph{\texttt{deepseek-v4-pro} under the adaptive policy prompt ($\Phi=0.96$).} Before the shift, the prefix calls only $B$. At $t=41$, the agent probes all three groups, finds $C$, and then uses $C$ alone.

\begin{center}\footnotesize
\setlength{\tabcolsep}{4pt}
\begin{tabular}{@{}c l p{0.48\columnwidth}@{}}
\toprule
turn & group mix & route-class \\
\midrule
36--40 & $\{1B\}$ & \texttt{current\_route} \\
\textbf{41} & $\{1A, 1B, 1C\}$ & \texttt{mixed\_route} (probe across groups) \\
42--50 & $\{1C\}$ & \texttt{current\_route} (locked onto reliable $C$) \\
\bottomrule
\end{tabular}
\end{center}

The boundary turn $t=41$ has six reasoning items.

\begin{examplebox}
\textbf{item 0 (initial):} \emph{Fresh request. ``Tuesday morning'' without ``this'' or ``next'' --- the upcoming Tuesday from Sunday June 21 is June 23, 2026. 75 minutes. Participants: Hiroshi and Lena.} \\
\textbf{calls:} \texttt{attendee\_preference\_lookup}, \texttt{timezone\_compare}, \texttt{room\_availability\_lookup}.
\medskip

\textbf{item 1 (after results):} \emph{Find slot.} \\
\textbf{calls:} \texttt{meeting\_slot\_generator}.
\medskip

\textbf{item 2 (after results):} \emph{\texttt{meeting\_slot\_generator} failed. Let me try \texttt{calendar\_lookup}.} \\
\textbf{calls:} \texttt{calendar\_lookup}.
\medskip

\textbf{item 3 (after results):} \emph{Both failed. Let me try another.} \\
\textbf{calls:} \texttt{slot\_window\_lookup}.
\medskip

\textbf{item 4 (after results):} \emph{\texttt{slot\_window\_lookup} worked. Tuesday 1:30 PM, Room 2A. Draft.} \\
\textbf{calls:} \texttt{concise\_email\_composer}.
\medskip

\textbf{item 5 (final):} \emph{Done.}
\end{examplebox}

Items 2--4 show immediate switching after failure. Within one turn, the model moves from $B$ (\texttt{meeting\_slot\_generator}) through $A$ (\texttt{calendar\_lookup}) to $C$ (\texttt{slot\_window\_lookup}).

\end{document}